\documentclass{article}
\usepackage{PRIMEarxiv}
\usepackage{makecell}
\usepackage[utf8]{inputenc}
\usepackage[T1]{fontenc}  
\usepackage{hyperref}      
\usepackage{breqn}
\usepackage{color}
\usepackage{url}
\usepackage{booktabs}

\usepackage[ruled,vlined,linesnumbered,algo2e]{algorithm2e}
\usepackage{algorithm}
\usepackage{algpseudocode}

\usepackage{amsmath,amssymb}
\usepackage{amsfonts}
\usepackage{nicefrac}    
\usepackage{microtype}   
\usepackage{lipsum}
\usepackage{fancyhdr}       
\usepackage{graphicx}       
\graphicspath{{media/}}     

\newcommand{\z}{\boldsymbol{z}}
\usepackage[subrefformat=parens]{subcaption}
\usepackage{comment}
\usepackage{cite}
\newcommand{\bx}{\boldsymbol{x}}
\newcommand{\bz}{\boldsymbol{z}}
\newcommand{\del}{\partial}
\usepackage{ulem}
\usepackage{bm}

\title{
	Physics-guided training of GAN to improve accuracy in airfoil design synthesis
}

\author{
  Kazunari Wada \\
  Department of Systems Innovation\\
Graduate School of Engineering\\
The University of Tokyo \\
Tokyo, JAPAN 113-8656\\
   \And
  Katsuyuki Suzuki \\
  Department of Systems Innovation\\
Graduate School of Engineering\\
The University of Tokyo \\
Tokyo, JAPAN 113-8656\\ 
  \AND
 Kazuo Yonekura \\
  Department of Systems Innovation\\
Graduate School of Engineering\\
The University of Tokyo \\
Tokyo, JAPAN 113-8656\\
}
\begin{document}
\maketitle

\begin{abstract}
Generative adversarial networks (GAN) have recently been used for a design synthesis of mechanical shapes. 
A GAN sometimes outputs physically unreasonable shapes. 
For example, when a GAN model is trained to output airfoil shapes that indicate required aerodynamic performance, significant errors occur in the performance values.
This is because the GAN model only considers data but does not consider the aerodynamic equations that lie under the data. 
This paper proposes the physics-guided training of the GAN model to guide the model to learn physical validity. 
Physical validity is computed using general‐purpose software located outside the neural network model. Such general-purpose software cannot be used in physics-informed neural network frameworks, because physical equations must be implemented inside the neural network models. 
Additionally, a limitation of generative models is that the output data are similar to the training data and cannot generate completely new shapes. 
However, because the proposed model is guided by a physical model and does not use a training dataset, it can generate completely new shapes. 
Numerical experiments show that the proposed model drastically improves the accuracy. 
Moreover, the output shapes differ from those of the training dataset but still satisfy the physical validity, overcoming the limitations of existing GAN models.
\end{abstract}


\section{Introduction}
The mechanical design aims to design shapes that satisfy performance requirements.
This task is referred to as an inverse design problem.
Traditionally, designers must perform trial and error to obtain desired shapes. 
However, this process is time-consuming and relies on the designer's skills. 
Inverse problems can be solved using mathematical optimization methods, such as sensitivity analysis \cite{sensitivity_analysis1, sensitivity_analysis2} and the adjoint method \cite{adjoint}; however, they require long computation times. 
Data analysis techniques such as principal component analysis  \cite{pca1, pca2} and surrogate models have also been employed to perform dimension reduction and model direct problems \cite{yonekura14}; however, these methods do not directly search for shapes based on the required performance.

In recent years, 
deep generative models such as the variational autoencoder (VAE) \cite{original VAE} 
and generative adversarial networks (GANs) \cite{original GAN} have been employed to solve engineering problems\cite{DGM-eng1, DGM-eng2} including airfoil designs. 
These generative models can generate data similar to training data by capturing their characteristics.
\cite{yonekura_vae,yonekura_vae2} used conditional VAE (CVAE) to generate shapes that closely match the specifications required.
\cite{cGAN_airfoil1,cGAN_airfoil2} employed a conditional GAN (cGAN) to design airfoils. 
A limitation of these methods is that the obtained shapes have zigzag lines, and fluid computations cannot be performed. 
To overcome the issue of zigzag lines, B\'{e}zier-GAN \cite{B-GAN}, which adds b\'{e}zier curves to the network, and BSplineGAN \cite{B-Spline}, which incorporates B-splines, have been proposed.
Lin et al. \cite{CST-GANs} proposed CST -GAN, which combines the CST method and GAN to optimize the parameters and generate smooth shapes. 
However, CST-GAN can only generate airfoils that can be represented by parameters, thereby limiting the variation in shape.
As a post-processing method for smoothing the shape, Wang et al. \cite{Airfoil GAN} applied a smoothing filter to the shape generated by cGAN.
However, because smoothing is not performed in the GAN model, there is no direct relationship between the conditional label and the performance values of the smoothed shapes. 
In addition, the performance values of the shapes before and after smoothing may not match. 

Smooth curves are essential in airfoil design for calculating the performance. 
However, there were issues owing to differences in performance between the generated and smoothed shapes, as well as limited shape variation. 
Yonekura et al. \cite{yonekura_wgangp} proposed a method using a combination of cGAN and Wasserstein GAN with a gradient penalty, called a conditional Wasserstein GAN with a gradient penalty (cWGAN-gp). This indicates that smooth airfoil shapes can be generated 
without smoothing. 
This method enables shape generation under specific conditions.
This study successfully overcame the issue of zigzag lines in the generated shapes. 
Nonetheless, the error in the performance value remains significant.
Because the learning model does not consider the aerodynamic equations, 
it is not possible to confirm whether the generated shapes are physically valid.

Physics-informed neural network (PINN) \cite{PINN_method1,PINN_method2} has been proposed as a machine-learning method that satisfies the governing 
equations. 
PINN has been widely adopted in numerous domains \cite{PINN1,PINN2,PINN3,PINN4,PINN5}. 
Daw et al. \cite{PINN_lake} successfully applied this method to predict lake temperature and improve the prediction accuracy. 
PINN has also been applied to inverse problems \cite{PINN-inv1,PINN-inv2,PINN-inv3,PINN-inv4}.
To perform the backpropagation of PINN, gradients that follow the governing equations are required, 
and the equations must be described within the neural networks. 
However, it is difficult to use external software, such as
general-purpose CAE software, because backpropagation cannot be performed. 
Therefore, the algorithm must be written inside neural network programs.
In other words, it is not feasible to include the output of external software, such as CAE software, in neural network models. This is an essential issue in engineering research.

To assess this issue, the concept of a physics-guided GAN (PG-GAN) was described \cite{yonekura_PG-GAN} using a simple fundamental problem: Newton's equations of motion. 
In contrast to the original GANs, PG-GAN classifies data as true or fake based on physical reasons and learns to generate physically valid outputs.
It enables arbitrary software to guide neural networks, but has not been studied for inverse design problems. 
In this study, based on the concept of PG-GANs, a PG-GAN method for inverse design problems is proposed. 
To compute the physical validity, we utilize a flow computation program that calculates the lift coefficient (performance value) for a shape using aerodynamic equations. In other words, our model learns by judging whether the performance values of the generated shapes satisfy software requirements.
A naive PG-GAN requires extensive computational time, which is not feasible in actual applications. The present study proposes an approximation method for the PG-GAN to reduce the computation time. In addition, zigzag lines are observed in the generated shapes. By quantifying and minimizing the degree of distortion in the model, this problem is overcome, and smooth shapes are generated.

The remainder of this paper is organized as follows. 
The GAN models are described in Section 2. The airfoil design formulation and physics-informed neural networks are introduced in Section 3. The PG-GAN model is presented in Section 4. Numerical experiments are discussed in Section 5.
Finally, the conclusions are presented in Section 6.

\section{GAN models}
\subsection{GAN and conditional GAN}
\begin{figure}[!b]
  \begin{minipage}[b]{0.49\linewidth}
    \centering
    \includegraphics[keepaspectratio, scale=0.5]{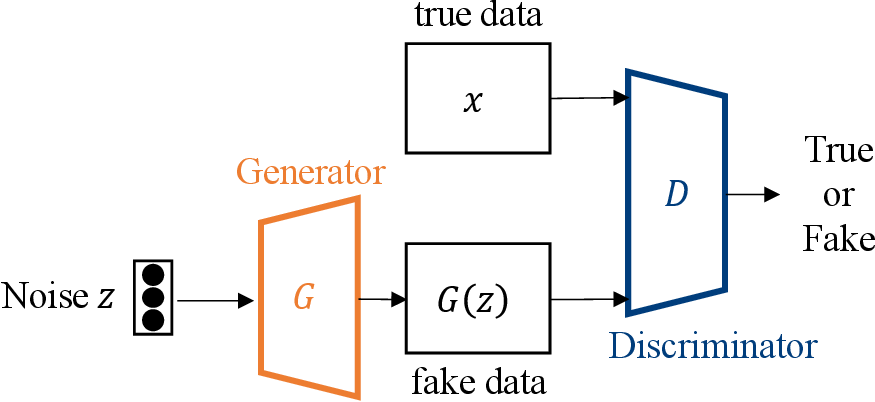}
    \subcaption{GAN.}
    \label{GAN_model}
  \end{minipage}
  \begin{minipage}[b]{0.49\linewidth}
    \centering
    \includegraphics[keepaspectratio, scale=0.5]{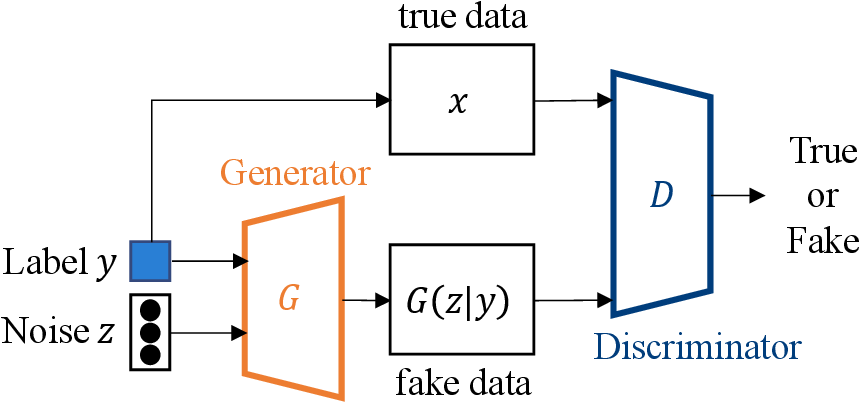}
    \subcaption{Conditional GAN.}
     \label{cGAN_model}
  \end{minipage}
  \caption{GAN and cGAN Architectures.}

\end{figure}

GAN \cite{original GAN} utilizes two networks: the generator $G$ and discriminator $D$. 
The generator is trained to generate fakes that mimic true data to deceive the discriminator. By contrast, the discriminator is trained to correctly discriminate between true and false data.
If both networks are trained and compete effectively, the generator can generate new data that resemble the true data. 
Generator $G$ produces fake data $G(\z)$ from random noise $\boldsymbol{z}$. 
The discriminator $D$ is trained to input true data $x$ and fake data $G(\z)$ and return 1 if the input data is true and 0 if it is not. The loss function is expressed as follows:
\begin{dmath}
\label{GAN_obj}
V(D,G)=\mathbb{E}_{\boldsymbol{x} \sim p_r(\boldsymbol{x})} \Big[\log(D(\boldsymbol{x}))\Big] 
	+ \mathbb{E}_{\boldsymbol{z}\sim p_z(\boldsymbol{z})} \Big[\log(1-D(G(\boldsymbol{z})))\Big],
\end{dmath}
where $p_r$ represents the true data distribution, and $p_{z}$ represents the noise probability distribution.
The discriminator $D$ is trained to decrease the value of 
$D(G(\boldsymbol{z}))$ and generator $G$ 
is trained to increase the value of $D(G(\boldsymbol{z}))$.
Hence, GAN is trained using $\min_G \max_D V(D, G)$.
The architecture of the GAN model is illustrated in Fig. \ref{GAN_model}.

To generate data under specific conditions, conditional GAN (cGAN) \cite{cGAN} was proposed (Fig. \ref{cGAN_model}).
Compared to GAN, conditional labels $\boldsymbol{y}$ are present in cGAN.
It is formulated as follows:
\begin{equation}
\begin{aligned}
    &\min_{\substack{G}}\max_{\substack{D}}V(D,G),\\
	&V(D,G)=\mathbb{E}_{\boldsymbol{x} \sim p_r(\boldsymbol{x})} \Big[\log(D(\boldsymbol{x}|\boldsymbol{y}))\Big] 
	+ \mathbb{E}_{\boldsymbol{z}\sim p_z(\boldsymbol{z})} \Big[\log(1-D(G(\boldsymbol{z}|\boldsymbol{y})))\Big].
\end{aligned}
\end{equation}
\subsection{WGAN-gp}
The original GAN is known to experience problems such as training instability, mode collapse, and vanishing gradients \cite{GAN-problems,GAN-problems2}. 
Arjovsky et al. \cite{WGAN} applied the Wasserstein distance to GAN and proposed the Wasserstein GAN (WGAN).
The Wasserstein distance is the distance between two probability distributions and represents the minimum cost required to move from one distribution to another.
For two probability distributions, $p_r$ and $p_g$, the Wasserstein distance can be expressed as follows:
\begin{equation}
\label{Wasserstein distance}
W(p_r, p_g)= \underset{\gamma \in \Pi(p_r, p_g)}{\text{inf}}\mathbb{E}_{(\boldsymbol{x},\boldsymbol{y})\sim \gamma}[\|\boldsymbol{x}-\boldsymbol{y}\|].
\end{equation}
Owing to Kantorovich-Rubinstein duality, the Wasserstein distance (Eq. \ref{Wasserstein distance}) can be expressed as the following for $K$-Lipschitz functions $f$.
\begin{equation}
\label{W_K_EQ}
    W(p_r, p_g)= \frac{1}{K} \underset{\|f\|_L\leq K}{\text{sup}}\mathbb{E}_{\boldsymbol{x}\sim p_r}f(\boldsymbol{x})-\mathbb{E}_{\boldsymbol{y}\sim p_g}f(\boldsymbol{y}).
\end{equation}
Denoting the parameters of function $f$ as $w\in \mathcal{W}$ and $f_w$, Eq. \ref{W_K_EQ} can be 
represented as follows:
\begin{equation}
    \label{w_pq_2}
    \underset{w\in \mathcal{W}}{\text{max}}\left[\mathbb{E}_{\boldsymbol{x}\sim p_r}\left[f_w(\boldsymbol{x})\right]-\mathbb{E}_{\boldsymbol{y}\sim p_g}\left[f_w(\boldsymbol{y})\right]\right].
\end{equation}

In the context of GAN, $p_r$ denotes the distribution of true data and $p_g$ represents the distribution of fake data. 
Both distributions are subject to the constraint of 1-Lipschitz continuity. 
If the generator can express the probability distribution of noise $p_z(\boldsymbol{z})$ as $g_\theta(\boldsymbol{z})$, then Eq. \ref{w_pq_2} can be written as follows:
\begin{equation}
\label{wgan-eq}
W(p_r, p_g)=\underset{w \in \mathcal{W}}{\text{max}}\left[\mathbb{E}_{\boldsymbol{x}\sim p_r}\left[f_w(\boldsymbol{x})\right]-\mathbb{E}_{\boldsymbol{z}\sim p_z}\left[f_w(g_\theta(\boldsymbol{z}))\right]\right],
\end{equation}
where $g_\theta$ represents the generator with parameter $\theta$, and $f_w$ corresponds to the discriminator with parameter $w$.
Enforcing the strict constraint of 1-Lipschitz continuity on $f$ in a neural network implementation is difficult, which has led to the proposal of weight clipping as an approximation method.
However,  the selection of clipping thresholds can result in training instability.

Gulrajani et al. \cite{WGAN-gp} incorporated a regularization term to ensure 1-Lipschitz continuity, instead of using on weight clipping.
The regularization term, called the gradient penalty, relaxes the 1-Lipschitz continuity constraint by imposing a penalty when the gradient deviates from 1. 
Hence, model WGAN-gp, which combines WGAN and gradient penalty, can be expressed as follows:
\begin{equation}
\begin{aligned}
&\min_{\substack{G}}\max_{\substack{D}}V(D,G),\\
    &V(D,G)=\mathbb{E}_{\boldsymbol{x}\sim p_r}\left[D(\boldsymbol{x})\right]-\mathbb{E}_{\boldsymbol{z}\sim p_z}\left[D(G(\boldsymbol{z}))\right]-\lambda \mathbb{E}_{\hat{\boldsymbol{x}}\sim p_{\hat{\boldsymbol{x}}}}\left[(\|\nabla_{\hat{\boldsymbol{x}}}D(\hat{\boldsymbol{x}})\|_2-1)^2\right],\label{cWGAN-gp-eq}
\end{aligned}
\end{equation}
where $\hat{\boldsymbol{x}}$ is the intermediate data obtained by linearly interpolating the data sampled from both the true and fake data distributions using random coefficients.

\section{Integrating physics models and neural networks}
\subsection{Airfoil design formulation}
By formulating the inverse problem, $\boldsymbol{x}$ satisfies the function $f$ and the parameter $\theta$ as
\begin{equation}
f(\boldsymbol{x};\theta)=0.
\end{equation}
In this case, determining $\boldsymbol{x}$ such that $f(\cdot;\bar{\theta})=0$ for a given parameter $\bar{\theta}$ is equivalent to obtaining a solution that satisfies the performance requirements. 
In the airfoil design problem, the shape $\boldsymbol{x}$ is represented by discretizing the corresponding curve on the $(x, y)$ plane into a set of 248 points, expressed as follows:
\begin{equation}
\boldsymbol{x}=(x_1,x_2,\cdots,x_{248},y_1,y_2,\cdots,y_{248})\in\mathbb{R}^{496}.
\end{equation}
To determine the lift coefficient $C_L$ of the airfoil $\boldsymbol{x}$, we use XFoil \cite{xfoil}.
The calculation for the airfoil $\boldsymbol{x}$ is denoted by a function $h$, such that
\begin{equation}
\label{eq_CL}
C_L=h(\boldsymbol{x}).
\end{equation}
With the function $h(\boldsymbol{x})$, the function $f(\boldsymbol{x};\theta)$ is represented as follows:
\begin{equation}
\label{eq_theta}
f(\boldsymbol{x};\theta)=\theta-h(\boldsymbol{x}).
\end{equation}
The airfoil design can be expressed as generating a shape $\boldsymbol{x}$ that satisfies $f(\boldsymbol{x};\bar{\theta})=0$, where $\bar{\theta}$ is the lift coefficient.
Let $G$ be a generator that takes a conditional label $\bar{c}$ and random noise $\boldsymbol{z}$ as inputs, and produces a shape $G(\boldsymbol{z}|\bar{c})$. 
When the performance value of the shape is denoted by $c$, Eq. \ref{eq_CL} can be written as follows:
\begin{equation}
\begin{aligned}
\label{eq-cs_CLR}
c&=h(\bx)\\
&=h(G(\boldsymbol{z}|\bar{c})).
\end{aligned}
\end{equation}

\subsection{Physics-informed neural networks}
Although machine learning can provide useful insights,
its output may not always conform to the governing equations. To overcome this limitation, PINNs \cite{PINN1,PINN2} has been developed. 
PINN is achieved by incorporating the governing equations that describe the physical system being trained as a part of the loss function of the neural network. 
This ensures that the network is trained to satisfy the physical constraints imposed by the problem.
For a neural network $\mathcal{F}$, let $\hat{Y}$ be its prediction and $Y$ be the variables for the governing equation. The loss function is given as follows:
\begin{equation}
\label{physics-loss}
    {\rm Loss} = {\rm Loss}(Y,\hat{Y}) + \lambda R(\mathcal{F}) +\gamma {\rm Loss}_{{\rm PHY}}(\hat{Y}),
\end{equation}
where $\lambda$ and $\gamma$ are hyperparameters.
The first and second terms correspond to the training error between the true value $Y$ and the predicted value $\hat{Y}$ and the regularization term that captures the complexity of the model.
These terms are common in machine learning models. By contrast, the third term is added to ensure physical consistency.
For example, assume that $Y$ satisfies the governing equation $\mathcal{G}(Y,Z)=0$ along with the other physical variables $Z$.
Here, by denoting ${\rm Loss}_{{\rm PHY}}(\hat{Y})$ as
\begin{equation}
    {\rm Loss}_{{\rm PHY}}(\hat{Y})=\| \mathcal{G}(\hat{Y},Z)\|_2^2,
\end{equation}
a physically informed model is constructed by computing $\arg\min_{\substack{\mathcal{F}}} {\rm Loss}$.

\begin{figure}[!t]
	\centering
	\includegraphics[width=150mm]{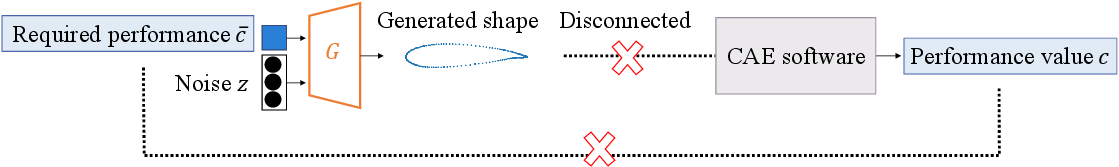}
		\caption{Relationship between generator $G$ and CAE software.}
		\label{cg_not_connected}
\end{figure}

In this study, using this PINN method to describe ${\rm Loss}_{{\rm PHY}}$ with the performance value $c=h(G(\boldsymbol{z}|\bar{c}))$,
the gradient regarding $c$ 
is expressed with the generator parameters $\phi_G$:
\begin{dmath}
\label{del_CLR}
\frac{\del c}{\del \phi_G}=\frac{\del h(G(\boldsymbol{z}|\bar{c}))}{\del \phi_G},
\end{dmath}
where the function $h$ signifies the CAE software calculation that determines the performance value of the input shape. 
However, the computation graph is disconnected between the generator and the CAE software (Fig. \ref{cg_not_connected}).
Consequently, if we represent the ${\rm Loss}_{{\rm PHY}}$ term in Eq. \ref{physics-loss} as ${\rm Loss}_{{\rm PHY}}(\bar{c},c)$ using $\bar{c}$ and $c$, the gradient $\frac{\del {\rm Loss}_{{\rm PHY}}(\bar{c},c)}{\del \phi_G}$ cannot be computed. In essence, it is not possible to directly utilize software outputs in the implementation of the PINN method.

\newpage
\section{PG-cWGAN-gp model for airfoil design}
\begin{figure}[!b]
	\centering
	\includegraphics[width=170mm]{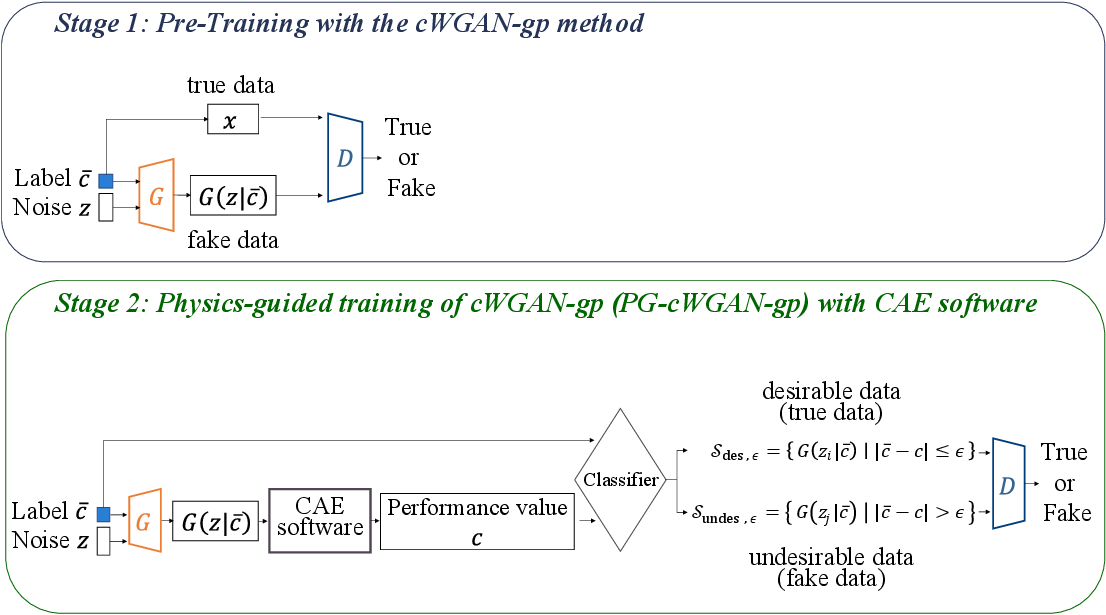}
		\caption{The proposed model.}
		\label{Proposed Model}
\end{figure}

We propose dividing the training process into two stages: Stage 1 for a pre-training and Stage 2 for the physics-guided training (Fig.\ref{Proposed Model}).
\begin{description}

	\item{Stage 1: } Pre-training with the cWGAN-gp method.
 \item{Stage 2: } Physics-guided training of cWGAN-gp (PG-cWGAN-gp) with the CAE software.
\end{description}

\subsection{Overview of the proposed model}

From Eq. \ref{eq_theta} and Eq. \ref{eq-cs_CLR}, assuming that the calculation operation by the CAE software is $h$, the performance value $c$ of the shape $G(z|\bar{c})$ can be written as $c=h(G(z|\bar{c}))$, where $\bar{c}$ is a conditional label. If the error between $\bar{c}$ and $c$ is sufficiently small, $G$ can generate a shape that satisfies the required performance.
Subsequently, the set $\chi(\epsilon)$  is used as the criterion for physical adequacy to classify the generated shape for desirable data  
(i.e., equivalent to true data) or undesirable data (i.e., equivalent to fake data). 

\begin{equation}
\begin{split}
\label{set-epsilon}
 \chi(\epsilon)&=\left\{ G(\boldsymbol{z}|\bar{c})\mid \left|f(G(\boldsymbol{z}|\bar{c});\bar{c})\right|\leq \epsilon  \right\}\\
  &=\left\{ G(\boldsymbol{z}|\bar{c})\mid \left|\bar{c}-h(G(z|\bar{c}))\right|\leq \epsilon \right\}\\
 &=\left\{ G(\boldsymbol{z}|\bar{c})\mid \left|\bar{c}-c\right|\leq \epsilon \right\},
\end{split}
\end{equation}
where $\epsilon$ is a threshold.
Let $\mathcal{S}_{{\rm des,} \epsilon}$ and $\mathcal{S}_{{\rm undes,} \epsilon}$ denote desirable and undesirable datasets, respectively. 
\begin{equation}
    \mathcal{S}_{{\rm des,} \epsilon}=\left\{G(z|\bar{c}) \mid \left|\bar{c}-c\right|\leq \epsilon   \right\}\, , \,   \mathcal{S}_{{\rm undes,} \epsilon}=\left\{G(z|\bar{c}) \mid  \left|\bar{c}-c\right|> \epsilon   \right\}.
\end{equation}

The loss function can be expressed in the cWGAN-gp framework as follows:
\begin{dmath}
\label{pcWGAN-obj}
V(D,G,\epsilon)
=\mathbb{E}_{G(\boldsymbol{z}|\bar{c})\in \mathcal{S}_{{\rm des,} \epsilon}}\left[D(\boldsymbol{x}|\bar{c})\right]-\mathbb{E}_{G(\boldsymbol{z}|\bar{c})\in \mathcal{S}_{{\rm undes,} \epsilon}}\left[D(G(\boldsymbol{z}|\bar{c}))\right]
-\lambda \mathbb{E}_{\hat{\boldsymbol{x}}\sim p_{\hat{\boldsymbol{x}}}}\left[ (\left\| \nabla_{\hat{\boldsymbol{x}}}D(\hat{\boldsymbol{x}}|\bar{c})\right\|_2-1)^2\right].
\end{dmath}
The optimization problem is $\min_G \max_D V(D,G,\epsilon)$.
The threshold value $\epsilon$ is initially determined by taking a large value and decreasing it as the learning progresses. This is because in the early learning phase, $G$ is not capable of generating the desired shape with a sufficiently small error.
Therefore, $\epsilon$ should be gradually reduced to train the model under stricter conditions. For $\epsilon > \epsilon^\prime$, the loss function $V(D, G,\epsilon)$ is used at the beginning of the training.
Under this loss function, $G$ learns to generate shapes that are similar to the desirable data, that is, shapes with $|\bar{c}-c| \leq \epsilon$.
After a certain number of epochs, the loss function is changed to $V(D, G,\epsilon^\prime)$. 
Consequently, desirable data are defined as shapes with $|\bar{c}-c| \leq \epsilon^\prime$, and $G$ learns to produce shapes similar to the desirable data.
Repeating this with $\epsilon \to 0$, the relationship between the conditional label $\bar{c}$ and the performance value $c$ is $|\bar{c}-c|\to 0$. 
Therefore, the generator can theoretically produce shapes with $\bar{c}=c$ at the end of the training process.

The loss function (Eq. \ref{pcWGAN-obj}) is optimized with respect to the label $\bar{c}$.
However, the airfoil dataset does not contain multiple shapes with the same lift coefficient. 
A unique feature of GAN is that, when generating data with the same label using the trained parameters, the generated shapes differ. 
In other words, given the same label $\bar{c}$ and two different noise vectors $\boldsymbol{z}_i$ and $\boldsymbol{z}_j$ $(i\neq j)$, the generated shapes are different. 
Thus, $G(\boldsymbol{z}_i|\bar{c})\neq G(\boldsymbol{z}_j|\bar{c})$. Consequently, their performance values differ: $h(G(\boldsymbol{z}_i|\bar{c}))\neq h(G(\boldsymbol{z}_j|\bar{c}))$.
Considering this property, we construct a conditional model that generates multiple shapes under the same label. 
In the implementation, multiple conditional labels $\bar{c}$ are selected at various points, 
which are called control labels $\boldsymbol{\bar{c}_{\rm ctrl}}$. From each control label $\bar{c}_{{\rm ctrl}_j}$ and random noise $\boldsymbol{z}$, numerous shapes are generated and used to construct a dataset by classifying the shapes.
\begin{figure}[tb]
	\centering
	\includegraphics[width=120mm]{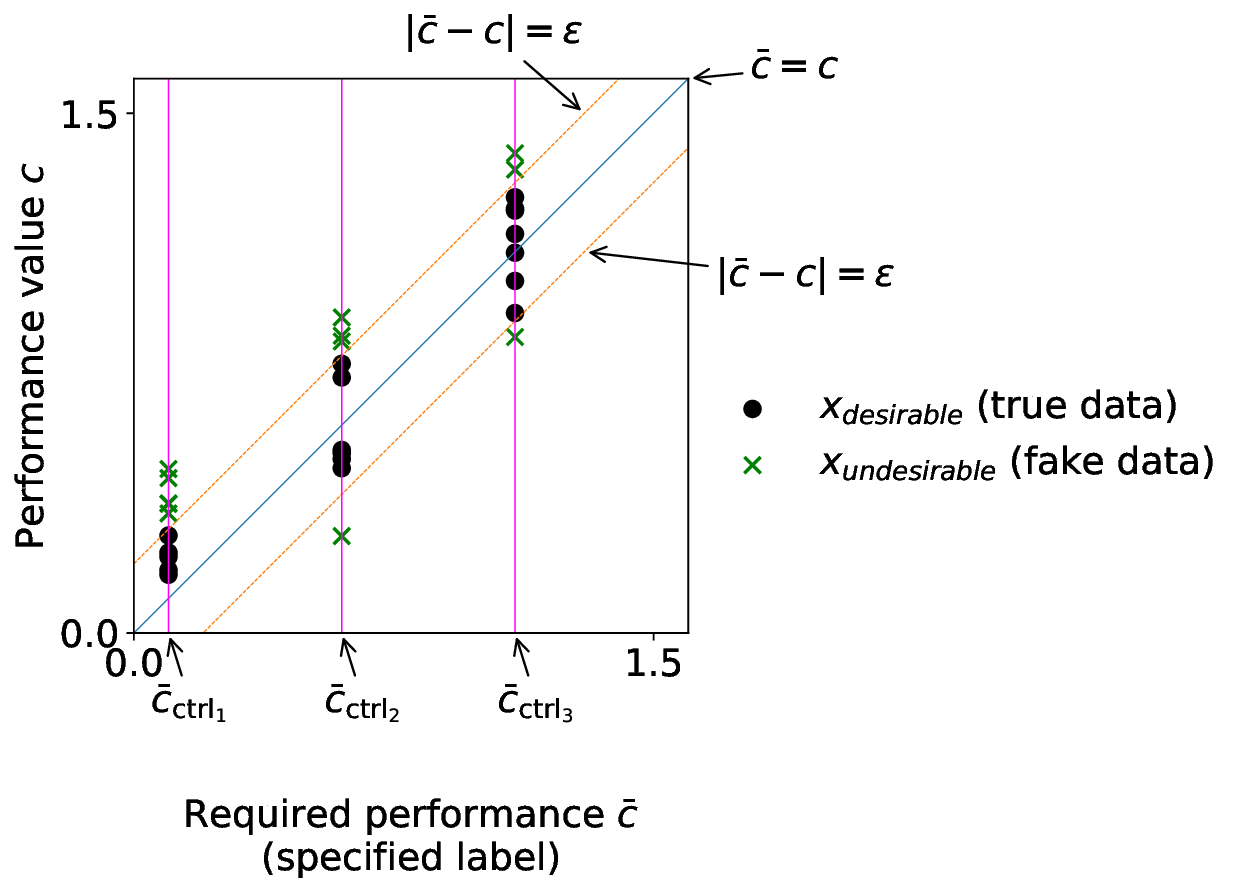}
		\caption{Classification of desirable and undesirable data, and setting of control labels.}
		\label{Classification of desirable and undesirable data}
\end{figure}
An overview diagram of the use of three control labels ($\bar{c}_{{\rm ctrl}_1}$, $\bar{c}_{{\rm ctrl}_2}$, $\bar{c}_{{\rm ctrl}_3}$) is shown in Fig. \ref{Classification of desirable and undesirable data}.
The horizontal axis denotes the conditional labels $\bar{c}$ (required performance) and the vertical axis represents the performance value of the generated shapes.
The loss function can then be rewritten as 
\begin{align}
\label{pcWGAN-rewritten}
   \begin{split}
&\mathbb{E}_{G(\boldsymbol{z}_i|\bar{c}_{{\rm ctrl}_j})\in \mathcal{S}_{{\rm des,} \epsilon}}\left[D(G(\boldsymbol{z}_i|\bar{c}_{{\rm ctrl}_j})|\bar{c}_{{\rm ctrl}_j})\right]-\mathbb{E}_{G(\boldsymbol{z}_j|\bar{c}_{{\rm ctrl}_j})\in\mathcal{S}_{{\rm undes,} \epsilon}}\left[D(G(\boldsymbol{z}_j|\bar{c}_{{\rm ctrl}_j}))\right]\\
    &\quad\quad-\lambda \mathbb{E}_{\hat{\boldsymbol{x}}\sim p_{\hat{\boldsymbol{x}}}}\left[ \left( \left\| \nabla_{\hat{\boldsymbol{x}}}D(\hat{\boldsymbol{x}}|\bar{c}_{{\rm ctrl}_j})\right\|_2-1 \right)^2\right]\\
    &=\mathbb{E}_{G(\boldsymbol{z}_i|\bar{c}_{{\rm ctrl}_j})\in \mathcal{S}_{{\rm des,} \epsilon}}\left[D(\boldsymbol{x}_{\rm desirable}|\bar{c}_{{\rm ctrl}_j})\right]-\mathbb{E}_{G(\boldsymbol{z}_j|\bar{c}_{{\rm ctrl}_j})\in \mathcal{S}_{{\rm undes,} \epsilon}}\left[D(\boldsymbol{x}_{\rm undesirable})\right]\\ 
    &\quad\quad-\lambda \mathbb{E}_{\hat{\boldsymbol{x}}\sim p_{\hat{\boldsymbol{x}}}}\left[ \left( \left\| \nabla_{\hat{\boldsymbol{x}}}D(\hat{\boldsymbol{x}}|\bar{c}_{{\rm ctrl}_j})\right\|_2-1 \right) ^2\right].
    \end{split}
\end{align}

\subsection{Pre-training}
We use the cWGAN-gp model proposed by \cite{yonekura_wgangp} for pre-training. 
The goal is to train the generator to produce smooth and diverse shapes 
from the random input noise $\boldsymbol{z}$ and the conditional labels $\bar{c}$. 
If the generator can produce shapes consisting of smooth curves, 
the generated shapes can be calculated for performance using the CAE software. 
The training algorithm is presented as Algorithm \ref{cWGAN-gp}. 
 \begin{algorithm}[!b]
\setcounter{AlgoLine}{0} 
\caption{Conditional WGAN with gradient penalty. Default values of $\lambda=10$, $n_{\rm critic}=5$, $\alpha=0.0001$, $\beta_1=0$, and $\beta_2=0.9$ were used.} 
\label{cWGAN-gp}
\SetAlgoLined
\SetKwInOut{Require}{Require}
\Require{The gradient penalty coefficient $\lambda$, 
the number of critic iterations per generator iteration $n_{\rm critic}$, the batch size $m$, and Adam hyperparameters $\alpha$, $\beta_1$, $\beta_2$.}
\Require{Initial critic parameters $w_0$ and initial generator parameters $\theta_0$.}
\While{$\theta$ has not converged}{
    \For{$t=1, \cdots, n_{\rm critic}$}{
        \For{$i=1, \cdots, m$}{
            Sample true data $\boldsymbol{x} \sim p_r$, true data labels $\boldsymbol{C}_L$, latent variables $\boldsymbol{z} \sim p_z(\boldsymbol{z})$, and a random number $\delta \sim U[0,1]$.
            
            $\tilde{\boldsymbol{x}} \leftarrow G_{\theta}(\boldsymbol{z} \,|\, \boldsymbol{C}_L)$
            
            $\hat{\boldsymbol{x}} \leftarrow \delta \boldsymbol{x} + (1-\delta) \tilde{\boldsymbol{x}}$
            
            $L^{(i)} \leftarrow D_w(\tilde{\boldsymbol{x}}) - D_w(\boldsymbol{x}) + \lambda(\|\nabla_{\hat{\boldsymbol{x}}}D_w(\hat{\boldsymbol{x}})\|_2 - 1)^2$
        }
        $w \leftarrow {\rm Adam} \left( \nabla_w \frac{1}{m} \sum_{i=1}^m L^{(i)}, w, \alpha, \beta_1, \beta_2 \right)$
    }
    Sample a batch of latent variables $\{\boldsymbol{z}^{(i)}\}_{i=1}^m \sim p_z(\boldsymbol{z})$.
    
    $\theta \leftarrow {\rm Adam} \left( \nabla_{\theta} \frac{1}{m} \sum_{i=1}^m (-D_w(G_{\theta}(\boldsymbol{z} \,|\, \boldsymbol{C}_L))), \theta, \alpha, \beta_1, \beta_2 \right)$\;
}
\end{algorithm}
The shape and performance values of the training dataset are true data, and the generated shapes $G(\boldsymbol{z}|\bar{c})$ are considered as fake data,
which is produced by using noise $\boldsymbol{z}$ and the performance $C_L$ as the conditional label $\bar{c}$. 
The gradients for the discriminator $D$ with parameters $w$ and the generator $G$ with parameters $\theta$ are calculated using the loss function (Eq. \ref{cWGAN-gp-eq}). 
These gradients are used to update the parameters.
After training, the generator is capable of generating pseudo-shapes for the dataset under any conditional label $\bar{c}$.
\begin{figure}[!b]
 \begin{minipage}[b]{0.32\linewidth}
  \centering
  \includegraphics[keepaspectratio, width=40mm]
  {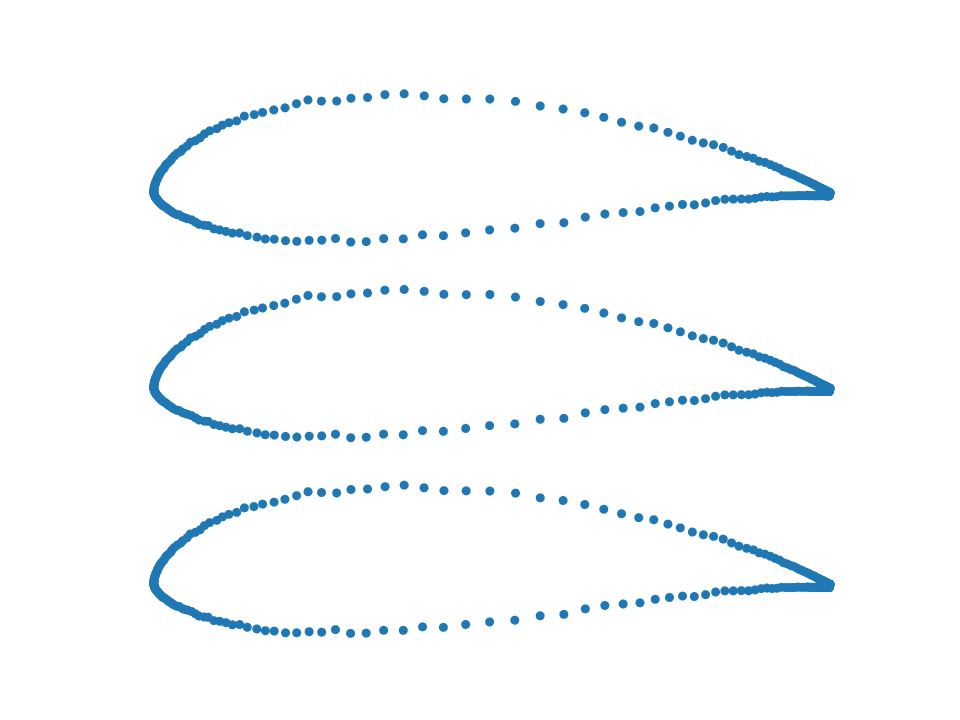}
  \subcaption{${\rm Epoch}=0$.}\label{sst12}
 \end{minipage}
 \begin{minipage}[b]{0.32\linewidth}
  \centering
  \includegraphics[keepaspectratio, width=40mm]
  {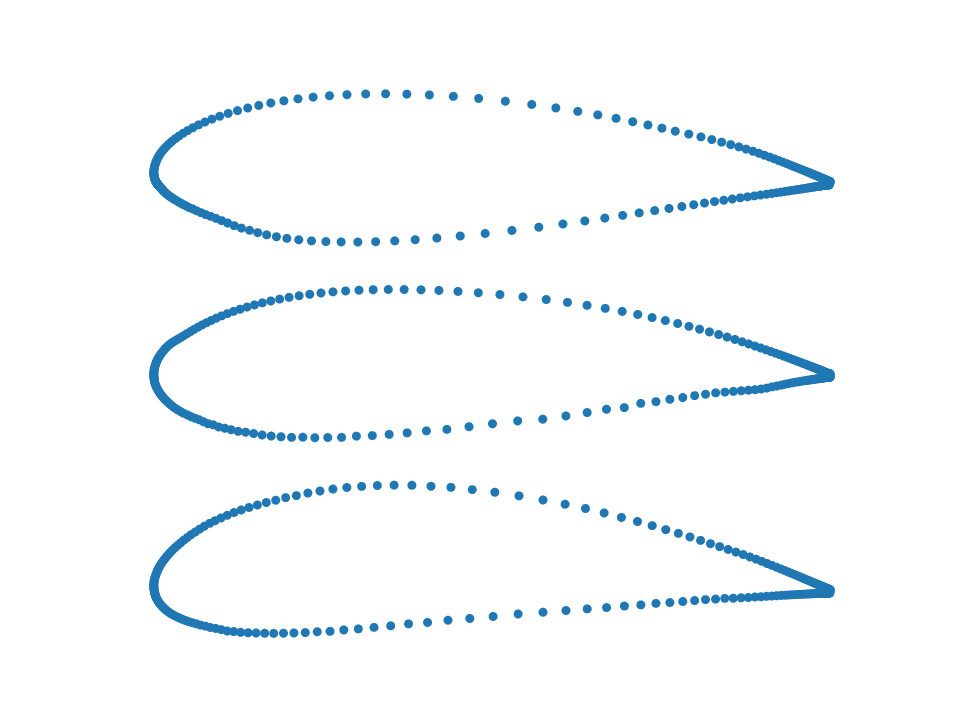}
  \subcaption{${\rm Epoch}=50000$.}\label{gst12}
 \end{minipage}
 \begin{minipage}[b]{0.32\linewidth}
  \centering
  \includegraphics[keepaspectratio, width=40mm]
  {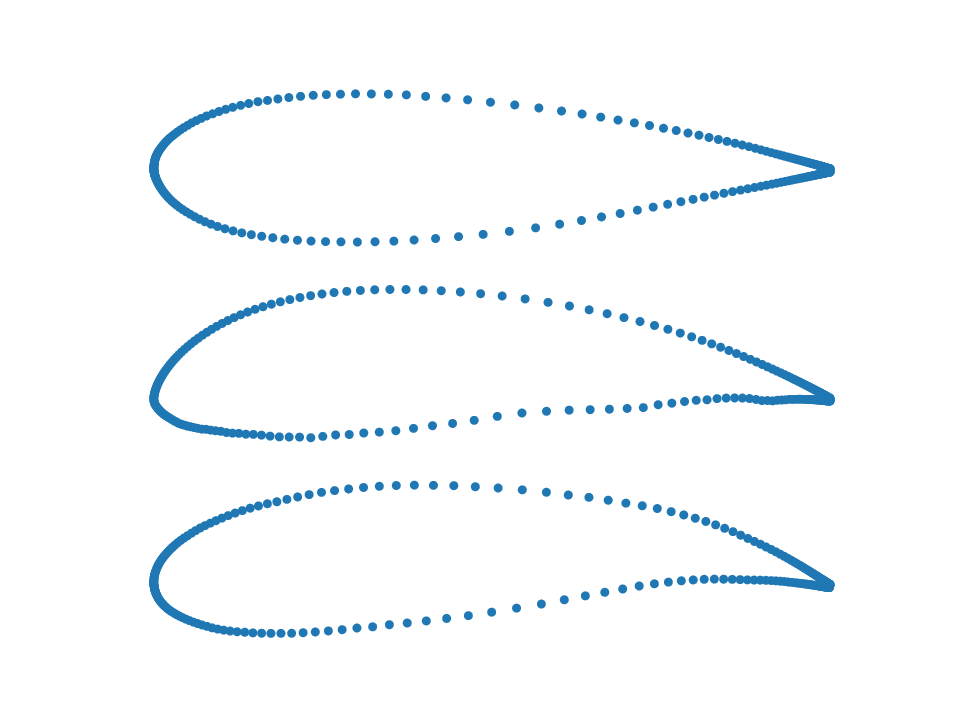}
  \subcaption{${\rm Epoch}=100000$.}\label{gic20}
 \end{minipage}
 \caption{Examples of generated shapes at each epoch.}\label{generated shapes at each epoch}
\end{figure}
\noindent
Fig. \ref{generated shapes at each epoch} displays examples of generated airfoil shapes at each epoch. 
At ${\rm Epoch}=0$, the shapes generated by the randomly initialized parameters of the generator $G$ are zigzagged. 
Consequently, performance calculations using CAE software do not converge, and performance values cannot be obtained. 
However, after training for ${\rm Epoch}=100000$, the generated shapes consist of smooth curves, 
and the shape performance value $c$ can be obtained through software calculations.

\subsection{PG-cWGAN-gp algorithm}

PG-cWGAN-gp is formulated and the algorithm is presented in Algorithm \ref{physics-guided conditional WGAN-gp}. The parameters for $G$ and $D$ are initialized using the pre-trained parameters $\theta_{\rm pretrained}$ and $w_{\rm pretrained}$. 
The CAE software is utilized within the network to ensure the physical validity of the generated shape.
In the PINN method, if the software output is directly described as a residual term, it is impossible to calculate gradients and incorporate them into the learning process. Hence, we create a loss function that takes advantage of the ability of the GAN model to generate pseudo-data for true data. The control labels are placed evenly in the interval $[0, 1.58]$ of $\bar{c}$.
The $\epsilon {\rm - } array$ is a set of values comprising the elements of $\epsilon$ in Eq. \ref{set-epsilon}. 
The elements of $\epsilon {\rm - } array$ are extracted and multiple control labels $\boldsymbol{\bar{c}_{\rm ctrl}}$ are selected at equal intervals in the range $[0,1.58]$ of $\bar{c}$. 
The $G$ attempts to generate a shape $\check{\bx}=G(\boldsymbol{z}|\boldsymbol{\bar{c}_{\rm ctrl}})$, where $\boldsymbol{\bar{c}_{\rm ctrl}}$ are the control labels and $\boldsymbol{z}$ is the random noise. 
Each generated shape is classified as desirable data or undesirable data. 
When the shape is desirable, it is added to $S_{\rm desirable}$ and its label is included in $Label_{\rm desirable}$. 
In the case where data are regarded as undesirable, the noise comprising its shape is added to $Z_{\rm undesirable}$, and its label is included in $Label_{\rm undesirable}$.

\begin{algorithm}[!b]
\caption{Physics-guided conditional WGAN with gradient penalty. Default values of $\lambda=10$, \,$n_{\rm critic}=5$,\, $\alpha=0.0001$, \,$\beta_1=0$, \,$\beta_2=0.9$, \, $m=10$, \, and $\epsilon {\rm -} array =(0.2, 0.1367, 0.0733, 0.01)$ are used.} 
\label{physics-guided conditional WGAN-gp}
  \setcounter{AlgoLine}{0} 
\SetAlgoLined
\SetKwInOut{Require}{Require}
\Require{The gradient penalty coefficient $\lambda$, the number of critic iterations per generator iteration
$n_{\rm critic}$, the batch size $m$, Adam hyperparameters $\alpha$, $\beta_1$, $\beta_2$, and the array of classification boundaries $\epsilon {\rm - } array$.}
\Require{Pretrained critic parameters $w_{\rm pretrained}$ and pretrained generator parameters $\theta_{\rm pretrained}$.}
$\theta \leftarrow \theta_{\rm pretrained}$, $w \leftarrow w_{\rm pretrained}$

\ForAll{$\epsilon \in \epsilon{\rm -} array$}{
    \While{$\theta$ has not converged}{
        \For{$t=1, \cdots, n_{\rm critic}$}{
            \For{$i=1, \cdots, m$}{
                Select control labels $\boldsymbol{\bar{c}_{\rm ctrl}}$ for specified label.

                Select latent variables $\boldsymbol{z}$.

 		$\check{\bx} \leftarrow G(\bz | \boldsymbol{\bar{c}_{\rm ctrl}})$

		$S_{\rm desirable} \leftarrow \phi: Set$, $Label_{\rm desirable} \leftarrow \phi: Set$

		$Z_{\rm undesirable} \leftarrow \phi: Set$, $Label_{\rm undesirable} \leftarrow \phi: Set$

                 \ForAll{$\check{\bx}_j \in \check{\bx}$,~ $\bar{c}_{{\rm ctrl}_j}\in \boldsymbol{\bar{c}_{\rm ctrl}}$, ~ $\boldsymbol{z}_j \in \boldsymbol{z}$}{
                    Compute the performance value $c$ of $\check{\boldsymbol{x}}_j$ using CAE software.

                    \If{$|\bar{c}_{{\rm ctrl}_j} - c| \leq \epsilon$}{
                        $S_{\rm desirable} \leftarrow S_{\rm desirable} \cup \{\check{\boldsymbol{x}}_j\}$

                        $Label_{\rm desirable} \leftarrow Label_{\rm desirable} \cup \{\bar{c}_{{\rm ctrl}_j}\}$
                    } \Else{
                        $Z_{\rm undesirable} \leftarrow Z_{\rm undesirable} \cup \{\boldsymbol{z}_j\}$

                        $Label_{\rm undesirable} \leftarrow Label_{\rm undesirable} \cup \{\bar{c}_{{\rm ctrl}_j}\}$
                    }
                }
                Sample desirable data $\boldsymbol{x} \in S_{\rm desirable}$, desirable data labels $\boldsymbol{y} \in Label_{\rm desirable}$, undesirable latent variables $\boldsymbol{\tilde{z}} \in Z_{\rm undesirable}$, undesirable data labels $\boldsymbol{\tilde{y}} \in Label_{\rm undesirable}$, and a random number $\delta \sim U[0,1]$.

                $\tilde{\boldsymbol{x}} \leftarrow G(\boldsymbol{\tilde{z}} \,|\, \boldsymbol{\tilde{y}})$

                $\hat{\boldsymbol{x}} \leftarrow \delta \boldsymbol{x} + (1-\delta) \tilde{\boldsymbol{x}}$

                $L^{(i)} \leftarrow D_w(\tilde{\boldsymbol{x}}) - D_w(\boldsymbol{x}) + \lambda(\|\nabla_{\hat{\boldsymbol{x}}}D_w(\hat{\boldsymbol{x}})\|_2 - 1)^2$\;
            }
            $w \leftarrow {\rm Adam} \left( \nabla_w \frac{1}{m} \sum_{i=1}^m L^{(i)}, w, \alpha, \beta_1, \beta_2 \right)$\;
        }
        $\theta \leftarrow {\rm Adam} \left( \nabla_{\theta} \frac{1}{m} \sum_{i=1}^m (-D_w(\tilde{\boldsymbol{x}})), \theta, \alpha, \beta_1, \beta_2 \right)$\;
    }
}
\end{algorithm}
\newpage
\subsection{Approximation algorithm of PG-cWGAN-gp}
The naive PG-cWGAN-gp requires a long computation time.
This is because the proposed model aims to classify each shape as a desirable data set based on the CAE calculations. In this algorithm, numerous shape generation and performance calculations are performed for each iteration.
Therefore, these calculations require considerable computational time, which is not suitable for real-world applications.

To overcome this issue, we introduce an approximation algorithm in which CAE calculations are performed only during the update of $\epsilon$ in Eq. \ref{pcWGAN-rewritten}. For clarity, the algorithm is described using two generators, $G_1$ and $G_2$, with parameters $\theta_{1}$ and $\theta_{2}$, respectively (Algorithm. \ref{approximation algorithm}).
\begin{algorithm}[!b] 
\setcounter{AlgoLine}{0} 
\caption{Approximation algorithm  of physics-guided conditional WGAN with gradient penalty. Default values of $\lambda=10$, \,$n_{\rm critic}=5$,\, $\alpha=0.0001$, \,$\beta_1=0$, \,$\beta_2=0.9$, \, $m=10$, and  $\epsilon {\rm -} array=(0.2, 0.1367, 0.0733, 0.01)$ are used.}
\label{approximation algorithm}
\SetAlgoLined
\SetKwInOut{Require}{Require}
\Require{The gradient penalty coefficient $\lambda$, the number of critic iterations per generator iteration
$n_{\rm critic}$, the batch size $m$, Adam hyperparameters $\alpha$, $\beta_1$, $\beta_2$, and the array of classification boundaries $\epsilon {\rm -} array$.}
\Require{Pretrained critic parameters $w_{\rm pretrained}$, pretrained generator parameters $\theta_{\rm pretrained}$, initial generator $G_1$ parameters $\theta_{1}$, and initial generator $G_2$ parameters $\theta_{2}$.}

$\theta_{1} \leftarrow \theta_{\rm pretrained},~ w \leftarrow w_{\rm pretrained}$\;

\ForAll{$\epsilon \in \epsilon{\rm -} array$}{
    Select control labels $\boldsymbol{\bar{c}_{\rm ctrl}}$ for specified label. 

    Select latent variables $\boldsymbol{z}$.

    $\check{\bx} \leftarrow G_{1_{\theta_{1}}}(\bz | \boldsymbol{\bar{c}_{\rm ctrl}})$

    $S_{\rm desirable} \leftarrow \phi: Set$, $Label_{\rm desirable} \leftarrow \phi: Set$

    $Z_{\rm undesirable} \leftarrow \phi: Set$, $Label_{\rm undesirable} \leftarrow \phi: Set$

    \ForAll{$\check{\bx}_j \in \check{\bx},~\bar{c}_{{\rm ctrl}_j}\in \boldsymbol{\bar{c}_{\rm ctrl}}, ~ \boldsymbol{z}_j \in \boldsymbol{z}$}{
    Compute the performance value $c$ of $\check{\bx}_j$ using CAE software.

    \If{$|\bar{c}_{{\rm ctrl}_j}-c| \leq \epsilon$}{
            $S_{\rm desirable} \leftarrow S_{\rm desirable} \cup \{\check{\bx}_j\}$

             $Label_{\rm desirable} \leftarrow Label_{\rm desirable} \cup \{\bar{c}_{{\rm ctrl}_j}$\}

        } \Else{

          $Z_{\rm undesirable} \leftarrow Z_{\rm undesirable} \cup \{\boldsymbol{z}_j\}$

          $Label_{\rm undesirable} \leftarrow Label_{\rm undesirable} \cup \{\bar{c}_{{\rm ctrl}_j}\}$

        }
    }

    $\theta_{2} \leftarrow \theta_{1}$

    \While{$\theta_{2}$ \rm{has not converged}}{
        \For{$t=1,\cdots, n_{\rm critic}$}
	{
            \For{$i=1,\cdots, m$}
	    {
                Sample desirable data $\boldsymbol{x} \in S_{\rm desirable}$, desirable data labels $\boldsymbol{y}\in Label_{\rm desirable}$, undesirable latent variables $\boldsymbol{\tilde{z}} \in Z_{\rm undesirable}$, undesirable data labels $\boldsymbol{\tilde{y}}\in Label_{\rm undesirable}$, and a random number $\delta \sim U[0,1]$.

                $\tilde{\bx} \leftarrow  G(\boldsymbol{\tilde{z}}|\boldsymbol{\tilde{y}})$

                $\hat{\bx} \leftarrow \delta \bx+(1-\delta)\tilde{\bx} $

                $L^{(i)} \leftarrow D_w(\tilde{\bx})-D_w(\bx)+\lambda(\|\nabla_{\hat{\bx}}D_w(\hat{\bx})\|_2-1)^2$

            }

           $w \leftarrow {\rm Adam} \left(\nabla_w \frac{1}{m} \sum_{i=1}^m L^{(i)}, w, \alpha, \beta_1, \beta_2 \right)$

        }

        $\theta_{2} \leftarrow {\rm Adam} \left( \nabla_{\theta} \frac{1}{m} \sum_{i=1}^m (-D_w(\tilde{\bx}))), \theta_{2}, \alpha, \beta_1, \beta_2 \right) $

    }

    $\theta_{1} \leftarrow \theta_{2}$
}
\end{algorithm}
\noindent Dataset is created by $G_1$ with a parameter $\theta_{1}$, which generates shapes $G_1(\boldsymbol{z}|\boldsymbol{\bar{c}_{\rm ctrl}})$ for control labels. 
The shapes are then classified based on their performance values $c=h(G_1(\boldsymbol{z}_j|\bar{c}_{{\rm ctrl}_j}))$, control labels $\bar{c}_{{\rm ctrl}_j}$, and the threshold $\epsilon$ from the $\epsilon${\rm -} array.
Training is performed using $G_2$ with parameters $\theta_{2}$. $D$ with a parameter $w$ is updated based on the gradients. 
After training with a predetermined number of epochs, the parameter $\theta_{2}$ for $G_2$ is assigned to the parameter $\theta_{1}$ for $G_1$.

In the original PG-cWGAN-gp algorithm, the datasets $\mathcal{S}_{{\rm des,} \epsilon}$ and $\mathcal{S}_{{\rm undes,} \epsilon}$ are created at each iteration using the CAE calculation for all generated shapes $G(\boldsymbol{z}|\boldsymbol{\bar{c}_{\rm ctrl}})$. Focusing on the generator, the  parameters $\theta$ are updated using their gradients. However, the approximation algorithm only creates datasets at predetermined times, and it iterates and trains the datasets once created during that time.
In summary, the datasets consist of generated shapes $G_{1_{\theta_{1}}}(\boldsymbol{z}|\boldsymbol{\bar{c}_{\rm ctrl}})$ with parameter $\theta_{1}$. Then, the $G_2$ is trained using these values, which are classified based on $\theta_{1}$. $\theta_{2}$ is updated every epoch, but is not reflected in the data set each time.


\subsection{Smoothing outline}
The shape generated by the PG-GAN is found to be distorted.
One way to reduce distortion in conditional shape generation is to perform post-processing smoothing. However, there is a large discrepancy between the required performance and the performance value after smoothing.
This is because a smoothing process is not performed inside the model.
First, we discuss the extent the post-processing smoothing used in previous studies deviates from the required performance. Subsequently, we describe a learning method to minimize the degree of distortion by quantitatively capturing it inside the model.

\begin{figure}[!b]
 \centering
  \begin{minipage}[b]{0.45\hsize}
    \centering
    \includegraphics[width=75mm]{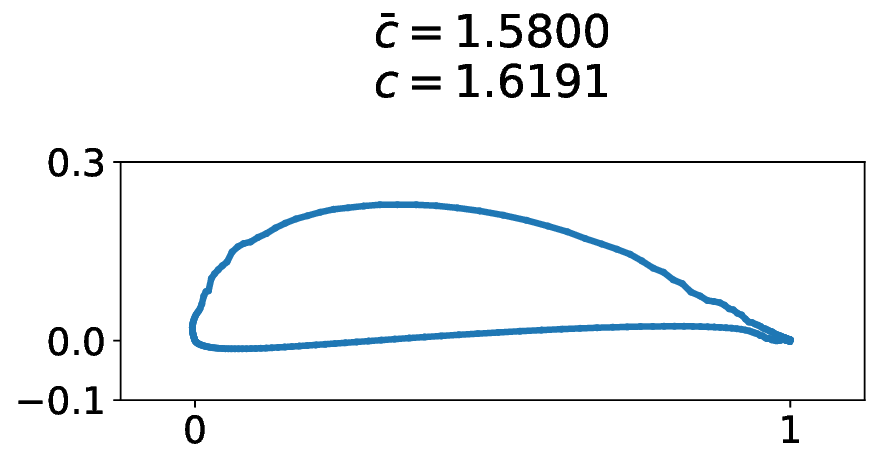}
    \subcaption{Before smoothing process.}
  \end{minipage}
  \begin{minipage}[b]{0.45\hsize}
    \centering
    \includegraphics[width=75mm]{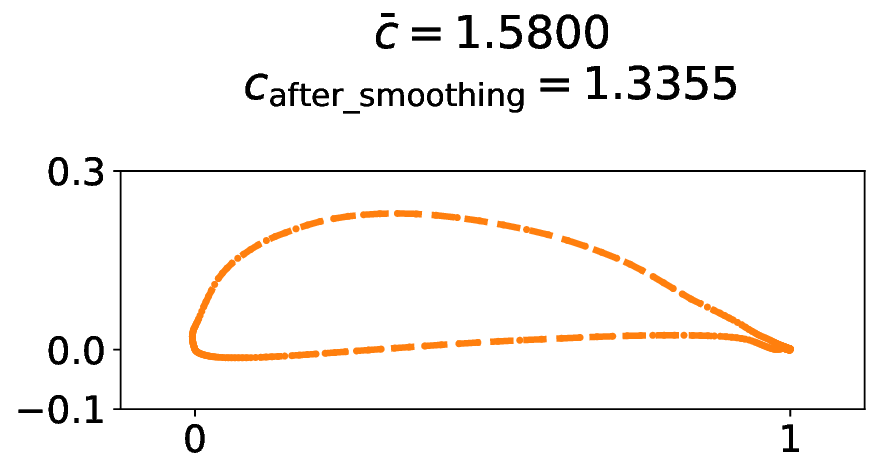}
    \subcaption{After smoothing process.}
  \end{minipage}
  \caption{Examples of shapes before and after smoothing.}
  \label{smoothing_process_bef_af}
\end{figure}

In a previous study, the Savitzky-Golay filter used for smoothing \cite{Airfoil GAN} was applied to distorted shapes. Using CAE calculations, the difference in the performance values of the shape with respect to the required performance before and after smoothing is shown in Fig. \ref{smoothing_process_bef_af}.
The performance value $c$ of the generated shape is $c=1.6191$ 
for the conditional label $\bar{c}=1.58$. However, the performance value 
of the shape after smoothing are $c_{{\rm after\_smoothing}}=1.3355$ and $c\neq c_{{\rm after\_smoothing}}$. 
It indicates that the performance values are not consistent before and after the smoothing process. 
Moreover, the performance value of the shape after smoothing is not related to the conditional label $\bar{c}$ because the smoothing process is outside the model.
Therefore, we formulated the degree of distortion and added it as a penalty term to the loss function within the generation model to  
generate smoothed shapes under the conditional labels.

\begin{figure}[!t]
 \centering
  \begin{minipage}[b]{0.45\hsize}
	\centering
	\includegraphics[width=60mm]{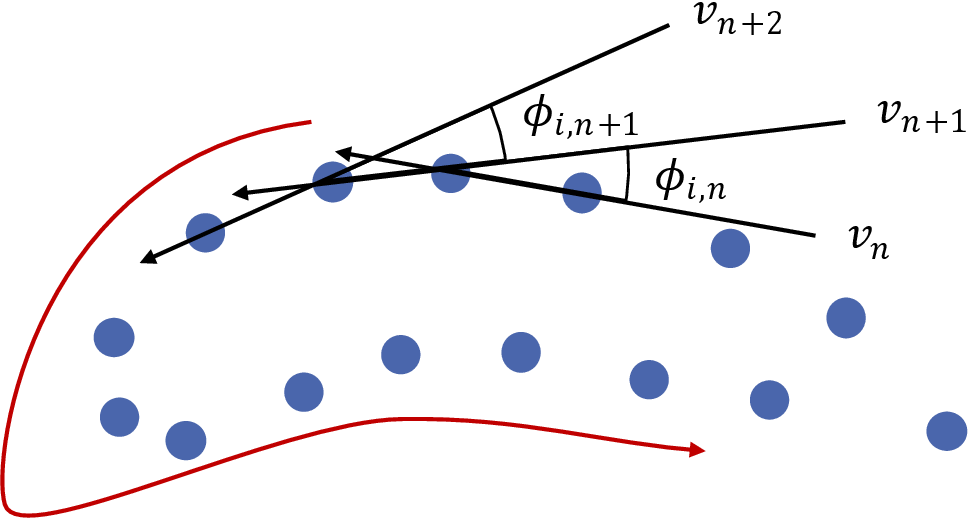}
 \subcaption{Shape with low distortion.}
  \end{minipage}
  \begin{minipage}[b]{0.45\hsize}
 	\centering
	\includegraphics[width=60mm]{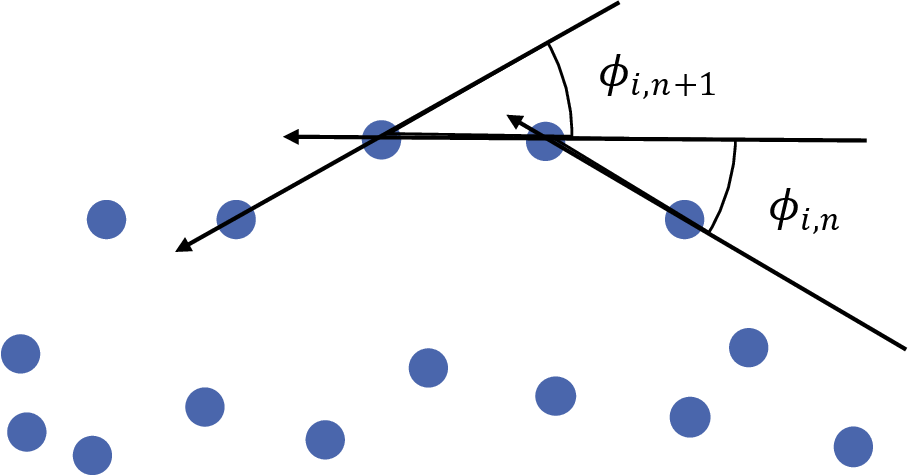}
    \subcaption{Shape with zigzag lines.}
  \end{minipage}
 \caption{Degree of distortion $\phi_{i,k}$ for a set of shape points $i$.}
    \label{phi-calculation} 
\end{figure}

The degree of distortion $\phi_i$ of a single shape point cloud $i$ is calculated using Eq. \ref{v} and Eq. \ref{phi_i}. 
The schematic diagram is shown in Fig. \ref{phi-calculation}.
\begin{equation}
\label{v}
\bm{v}_k=(\boldsymbol{x}_{k+1}-\boldsymbol{x}_k,\bm{y}_{k+1}-\bm{y}_k)^\top \qquad  (k=1,2,\cdots,N),
\end{equation}

\begin{dmath}
\label{phi_i}
\phi_i=\sum_{k=1}^N \phi_{i,k}
=\sum_{k=1}^N \arccos \left( \frac{\bm{v}_k^\top \bm{v}_{k+1} }{\|\bm{v}_k\| \|\bm{v}_{k+1}\|}  \right).
\end{dmath}

When considering distortion-free convex shapes such as circular shapes,
the degree of distortion $\phi_i$ 
is $\phi_i=2\pi$. 
By contrast, the shapes with higher distortions result in larger values of $\phi_i$, where $\phi_i\geq 2\pi$. Thus, by integrating $\phi_i$ into the model to minimize it, 
a reduction in distortion can be achieved. 
The distortion degree $\phi$ for multiple shapes is represented as the average of the distortion degrees $\phi_i$ for each shape point group $i$:
\begin{equation}
\phi=\frac{1}{l}\sum_{i=1}^l \phi_i.
\end{equation}
$\phi$ is introduced as a penalty term in the generator loss function.
In other words, the loss function for the generator $G$ in Eq. \ref{pcWGAN-rewritten} is denoted as $V_G$. By defining $\phi$ as $\phi(G(\boldsymbol{z}_j|\bar{c}_{{\rm ctrl}_j}))$, we can write $V_G$ as
\begin{equation}
\label{V_G}
V_G=
-\mathbb{E}_{G(\boldsymbol{z}_j|\bar{c}_{{\rm ctrl}_j})\in \mathcal{S}_{{\rm undes,} \epsilon}}\left[D(G(\boldsymbol{z}_j|\bar{c}_{{\rm ctrl}_j}))\right]
+\lambda_{\rm phi} \phi(G(\boldsymbol{z}_j|\bar{c}_{{\rm ctrl}_j})),
\end{equation}
where $\lambda_{\rm phi}$ denotes a hyperparameter.
Thus, the generator solves the minimization problem $\min_G V_G$.

\section{Numerical experiments}
\subsection{Training dataset and experimental conditions}

\begin{figure}[!b]
  \begin{minipage}[b]{0.49\linewidth}
    \centering
    \includegraphics[width=85mm]
    {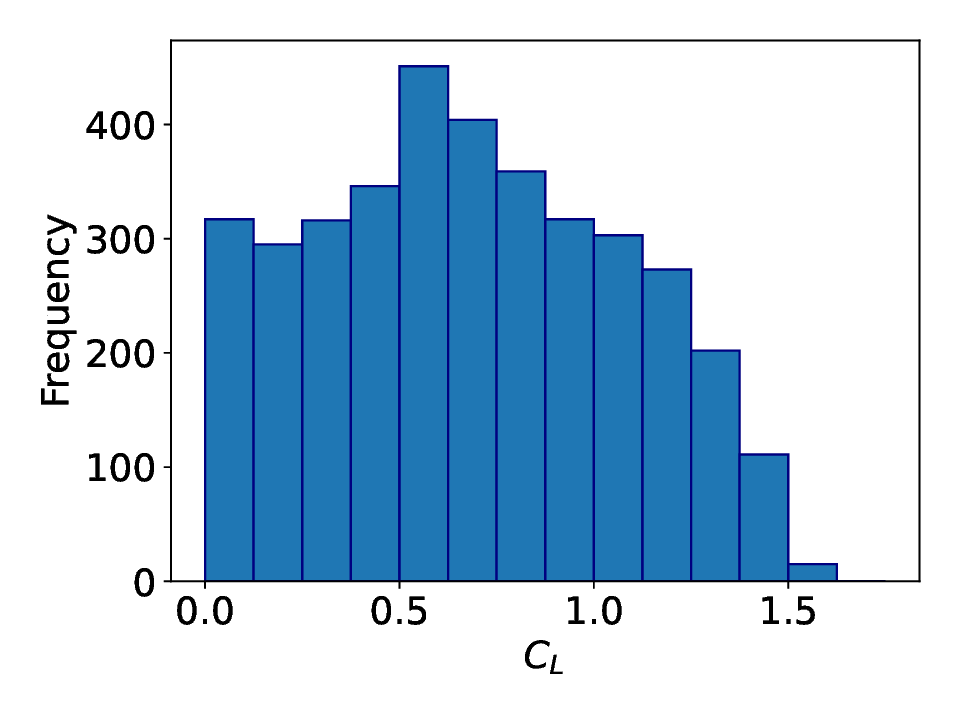}
    \subcaption{Lift coefficients and their frequency.}
  \end{minipage}
  \begin{minipage}[b]{0.49\linewidth}
    \centering
    \includegraphics[width=85mm]
    {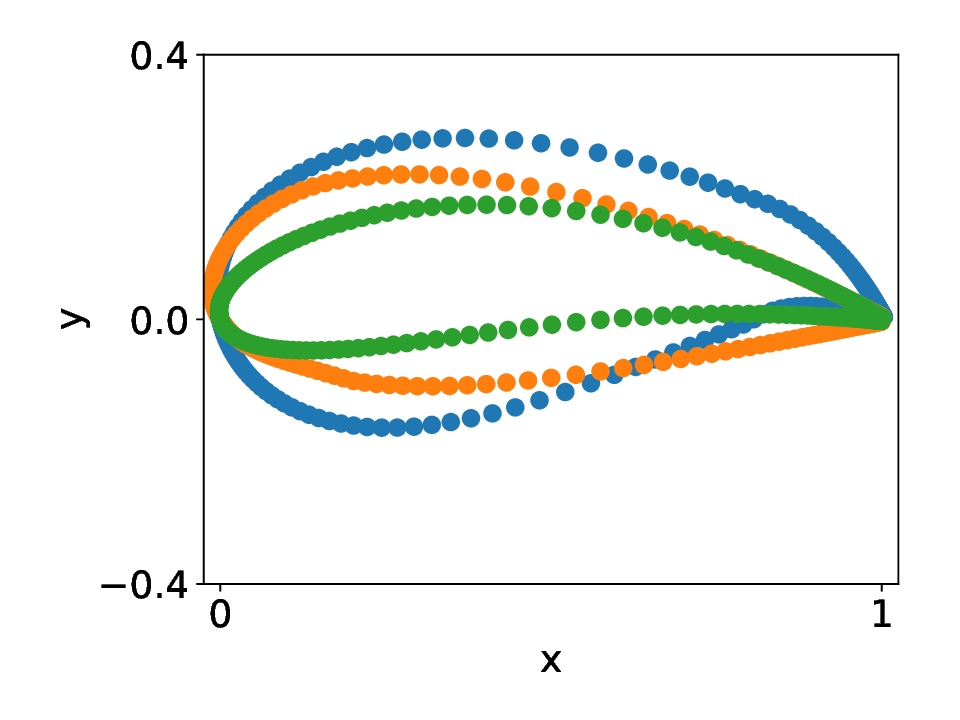}
    \subcaption{Examples of shapes.}
  \end{minipage}
  \caption{Overview of the dataset.}
  \label{Overview of the dataset}
\end{figure}

The training dataset is NACA airfoils from the four-digit series, as defined by the National Advisory Committee for Aeronautics (NACA).
In the four-digit series airfoils, the numerical value of each digit represents the values of the parameters of the airfoil in the following order: maximum camber $m$, location of maximum camber $p$, and thickness ratio $\tau$.
We use the NACA airfoils within the ranges of $m\in \{ 0, 1, \dots, 9\}$, $p\in \{ 0, 1, \dots, 9\}$, and $\tau \in \{ 0, 1, \dots, 99\}$.
Xfoil \cite{xfoil} was used to compute the flow field.
We set the angle of attack $\alpha$  to 5 degrees, Reynolds number ${\rm Re}$ to $3.0\times10^6$, and the maximum computational iterations to 100. 
Then, 
A total of 3709 data sets are used, excluding those for which $C_L$ could not be calculated using XFoil, $C_L<0$ or $C_L>2.0$. Each airfoil shape was then discretized into a sequence of 248 points on the $(x, y)$ plane corresponding to its shape (Fig. \ref{Overview of the dataset}) .

\subsection{Preliminary test comparing the exact and approximation algorithms}
\label{Preliminary_test}

The original PG-cWGAN-gp and the approximation algorithms are compared to verify their usefulness. The former is described as the exact algorithm, and the latter as the approximation algorithm.

\begin{figure}[!ht]
\begin{minipage}[b]{0.49\linewidth}
	\raisebox{.125\height}{\centering
	\includegraphics[width=85mm]{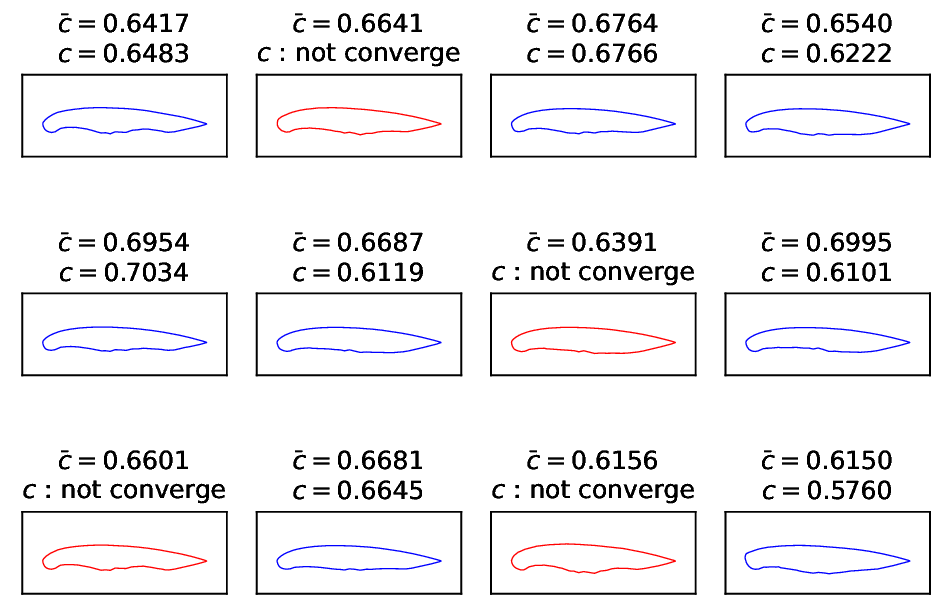}}
		\subcaption{Generated shapes.}
  \label{generated_shapes_by_exact_alg}
  \end{minipage}
  \begin{minipage}[b]{0.49\linewidth}
	\centering
	\includegraphics[width=70mm]{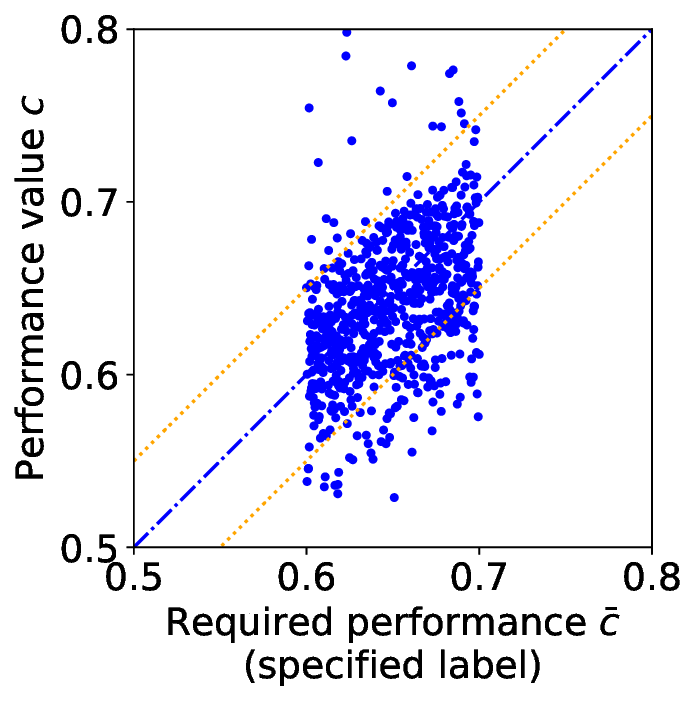}
		\subcaption{Scatter plot of performance value $c$ with conditional label $\bar{c}$.}
  \label{scatter_by_exact_alg}
  \end{minipage}
  \caption{Results from the exact algorithm.}
  \label{results_by_exact_alg}
\end{figure}

\begin{figure}[!ht]
\begin{minipage}[b]{0.49\linewidth}
	\raisebox{.125\height}{\centering
        \includegraphics[width=85mm]
        {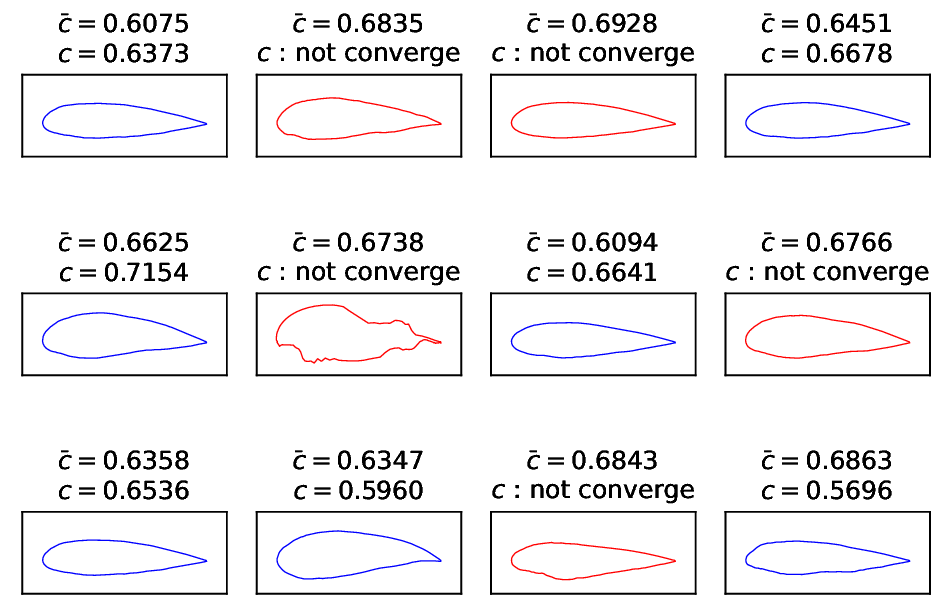}}
		\subcaption{Generated shapes.}
  \label{generated_shapes_by_approximate_alg}
  \end{minipage}
  \begin{minipage}[b]{0.49\linewidth}
	\centering
	\includegraphics[width=70mm]{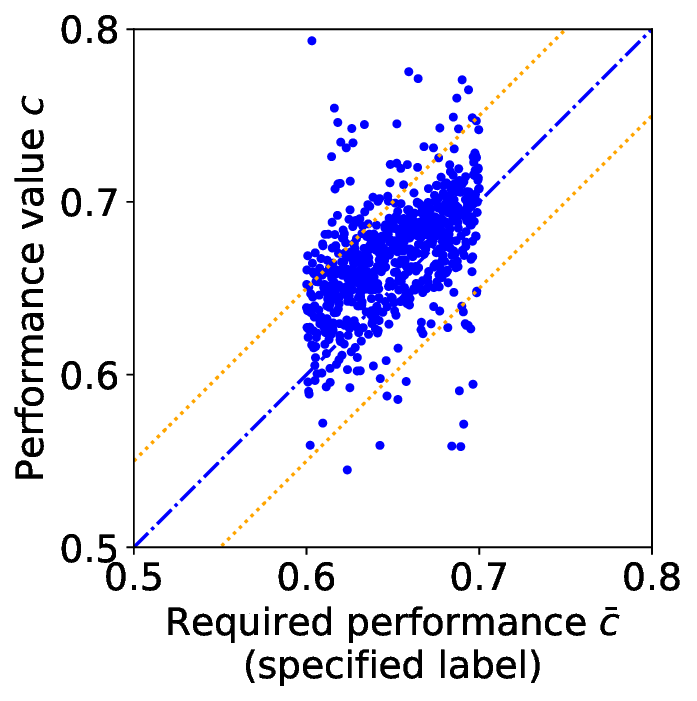}
		\subcaption{Scatter plot.}
  \label{scatter_by_approximate_alg}
  \end{minipage}
  \caption{Results from the approximation algorithm.}
  \label{results_by_approximate_alg}
\end{figure}

The two control labels, $\boldsymbol{\bar{c}_{\rm ctrl}}$ are set to 0.6 and 0.7 in the training process. 
Fig. \ref{results_by_exact_alg} and Fig. \ref{results_by_approximate_alg} show the shapes generated using the generator after learning when the required performance $\bar{c}$ is randomly specified between 0.6 and 0.7, and scatter plots of the performance values for the required performance. 
Both algorithms generate accurate shapes with small errors.
The blue shapes represent those for which the performance calculations of XFoil converge. 
The red shapes represent those for which the calculation did not converge (Fig. \ref{generated_shapes_by_exact_alg} and Fig. \ref{generated_shapes_by_approximate_alg}).
Some output shapes are different from the training dataset but still satisfy the requirements. 
In the ordinal machine learning method, such outputs that are different from the training cannot be obtained. 
Because the proposed method no longer uses a training dataset, the output differs from the data. 

The performance metric for the model was determined using both the MAE calculated using $\frac{1}{N}\sum_{i=1}^N |\bar{c}_i-c_{i}|$ and the XFoil results. 
Categorization of the generated shapes into successes or failures was also performed 
where shapes satisfying the condition $|\bar{c}-c|\leq0.05$ were labeled as successes, whereas those satisfying $|\bar{c}-c|>0.05$ were labeled as failures (Table \ref{Comparison of Models}).

\begin{table}[!t]
\centering
\caption{Comparison of algorithms when using two control labels.}
\label{Comparison of Models}
\begin{tabular}{lrrrrrr} \hline
& \multicolumn{1}{c}{\makecell[tc]{epoch in \\fine-tuning}} &  \multicolumn{1}{c}{success $\uparrow$} &  \multicolumn{1}{c}{failure $\downarrow$}& \multicolumn{1}{c}{not converge $\downarrow$}& \multicolumn{1}{c}{ MAE $\downarrow$ }&  \multicolumn{1}{c}{training time}\\ \hline 
cWGAN-gp \cite{yonekura_wgangp} & \multicolumn{1}{c}{-} & 1.0 $\%$ & 94.0 $\%$ & 5.0 $\%$  &0.2110& 10 hours\vspace{0.2mm}\\
\makecell[tl]{PG-cWGAN-gp \\ with exact algorithm} & \raisebox{-0.5em}{40} & \textbf{\raisebox{-0.5em}{69.4 $\%$} }& \textbf{\raisebox{-0.5em}{10.1 $\%$}}&\textbf{\raisebox{-0.5em}{20.5 $\%$}}  & \textbf{\raisebox{-0.5em}{0.0267}} & \raisebox{-0.5em}{10 hours + 8 hours}\vspace{0.2mm}\\
\makecell[tl]{PG-cWGAN-gp \\ with approximation algorithm} & \raisebox{-0.5em}{20000} & \raisebox{-0.5em}{61.8 $\%$} & \raisebox{-0.5em}{15.4 $\%$} & \raisebox{-0.5em}{22.8 $\%$}  &\raisebox{-0.5em}{0.0282} & \raisebox{-0.5em}{10 hours + 40 min.}\\ \hline
\end{tabular}
\end{table}

In both the exact and approximation algorithms, the amount of data for which $|\bar{c}-c|\leq0.05$ increases significantly. 
It implies that the generated shapes are more likely to satisfy the conditional label (required performance) than in the conventional method. 
When comparing the results of the exact and the approximate algorithms, 
the exact algorithm yields better results with respect to success, failure, nonconvergence rate, and MAE.
The computation time of the approximation algorithm is significantly less than that of the exact algorithm.

\subsection{Smoothing constraints}
Preliminary experiments show that the approximation algorithm can improve the accuracy while reducing the computation time.
Therefore, we expand the selection range of the control labels to $[0.01,1.58]$ and conduct an experiment using the approximation algorithm. We added the shape-smoothing constraints to Eq. \ref{V_G} and tested its effect.

\begin{figure}[!b]
\centering
\includegraphics[width=112mm]{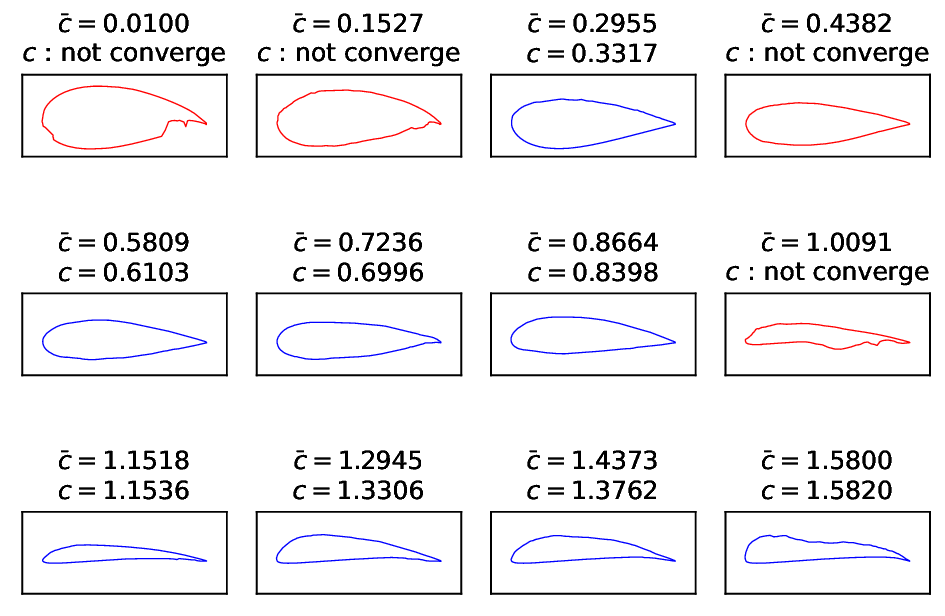}%
\caption{Generated shapes by uniform intervals $\bar{c}$ 
(PG-cWGAN-gp without smoothing constraints).}
\label{shapes_approxi_0_158_continuous}
\end{figure}

\begin{figure}[!t]
\centering
\includegraphics[width=112mm]{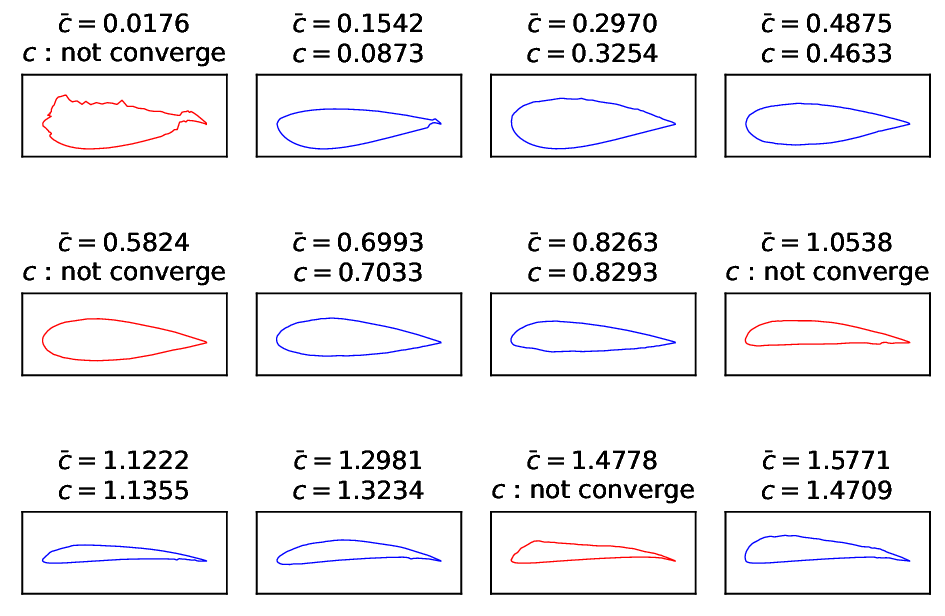}%
\caption{Generated shapes by random $\bar{c}$ (PG-cWGAN-gp without smoothing constraints).}
\label{shapes_approxi_0_158_random}
\end{figure}

\begin{figure}[!t]
    \centering
    
    \begin{subfigure}[b]{0.3\textwidth}
        \centering
        \includegraphics[width=\textwidth]{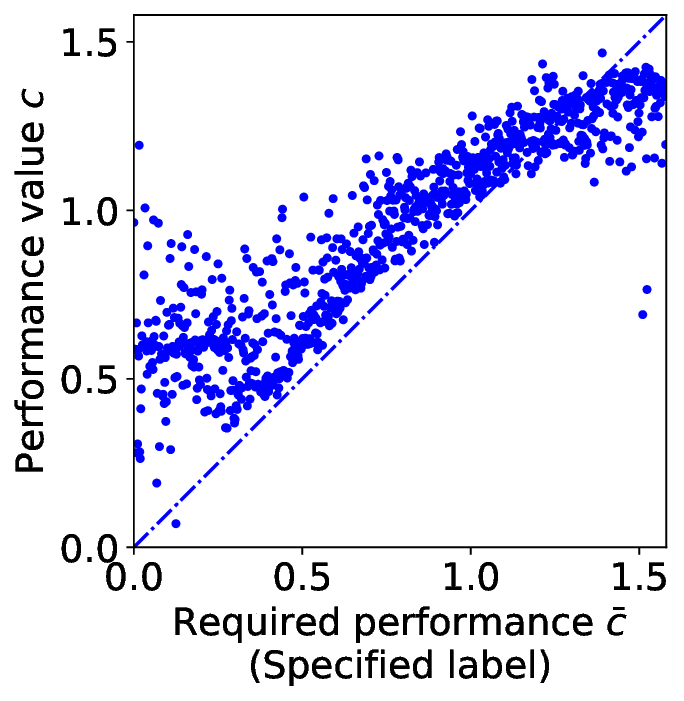}
         \caption{Scatter plot at the initiation.}
    \end{subfigure}
    \hfill
    \begin{subfigure}[b]{0.3\textwidth}
        \centering
        \includegraphics[width=\textwidth]{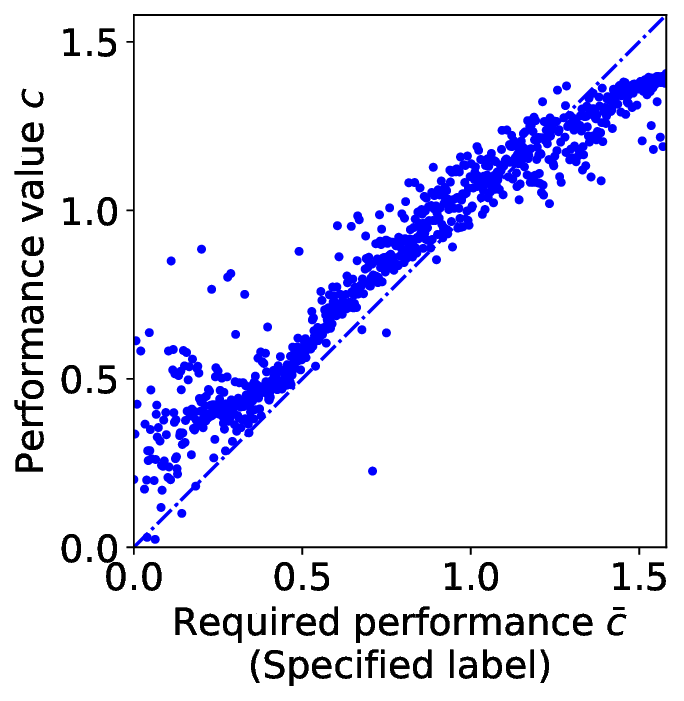}
        \caption{Scatter plot at 5000 epoch.}
    \end{subfigure}
    \hfill
    \begin{subfigure}[b]{0.3\textwidth}
        \centering
        \includegraphics[width=\textwidth]{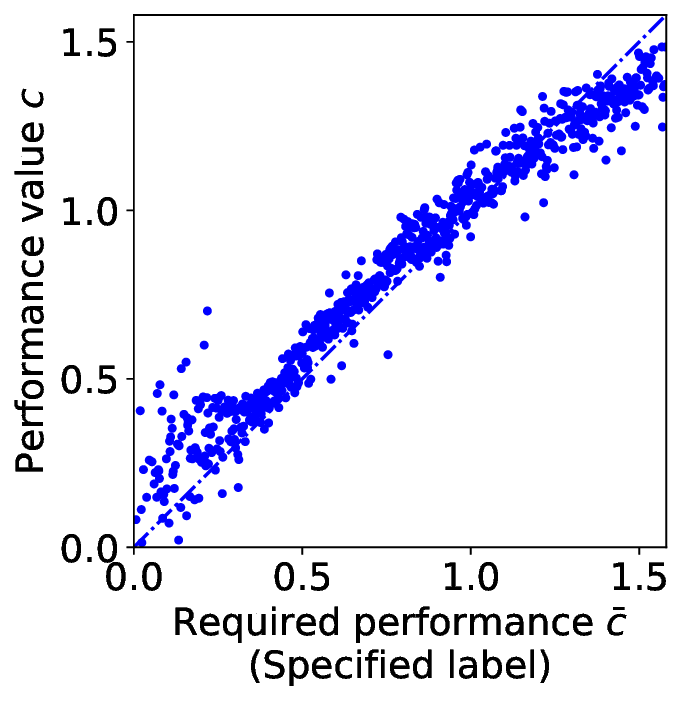}
        \caption{Scatter plot at 10000 epoch.}
    \end{subfigure}
    
    \vspace{1em}
    
    \begin{subfigure}[b]{0.3\textwidth}
        \centering
        \includegraphics[width=\textwidth]{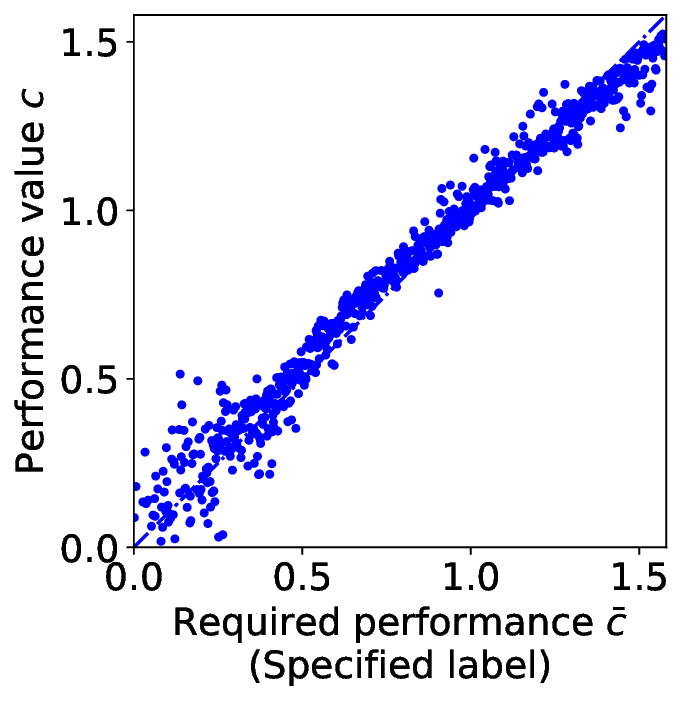}
        \caption{Scatter plot at 15000 epoch.}
    \end{subfigure}
    \hspace{0.05\textwidth}
    \begin{subfigure}[b]{0.3\textwidth}
        \centering
        \includegraphics[width=\textwidth]{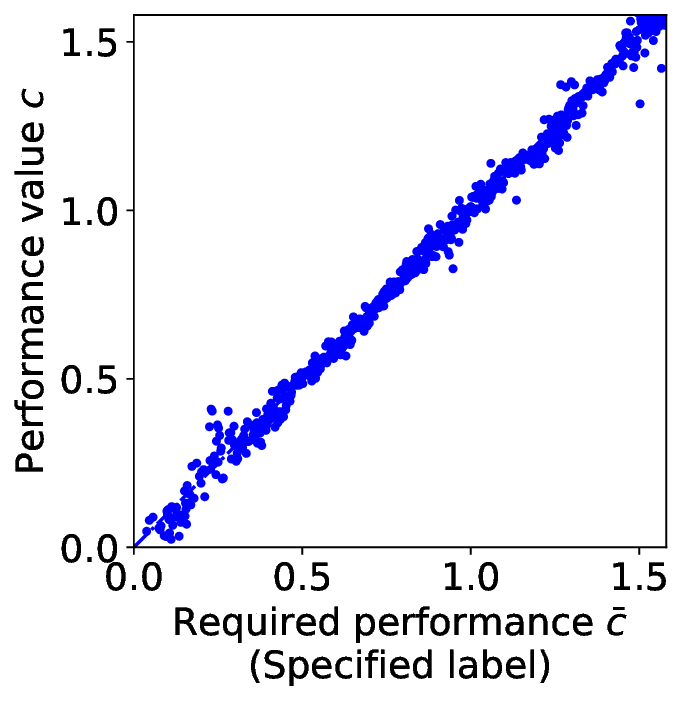}
        \caption{Scatter plot at 20000 epoch.}
    \end{subfigure}
    
  \caption{Scatter plot transition of PG-cWGAN-gp.}
  \label{fig:five_no_phi_scatters}
\end{figure}

Eight equidistant control labels, $\boldsymbol{\bar{c}_{\rm ctrl}}$ were selected from 0.01 to 1.58 and used for learning. 
The shapes produced by the generator after training are shown in Fig. \ref{shapes_approxi_0_158_continuous} and Fig. \ref{shapes_approxi_0_158_random}.
Fig. \ref{shapes_approxi_0_158_continuous} shows the shapes generated by selecting the required performance $\bar{c}$ from the interval $[0.01, 1.58]$ at uniform intervals, and Fig. \ref{shapes_approxi_0_158_random} shows the shapes with randomly chosen values of $\bar{c}$ from that interval.
Several shapes consist of zigzag lines, thereby indicating that the XFoil calculation is not converging.
Fig.\ref{fig:five_no_phi_scatters} provides the scatter plots of the required performance $\bar{c}$ and the performance value $c$ by 5000 epochs.
From the changes in the five scatter plots, the points are gradually plotted near the line $\bar{c}=c$. In other words, a shape with a performance value with a small error relative to the required performance is generated.
These results show that, for any of the required performances, the use of the approximation algorithm produces more accurate results than prior studies.

\begin{figure}[!t]
\centering
\includegraphics[width=112mm]{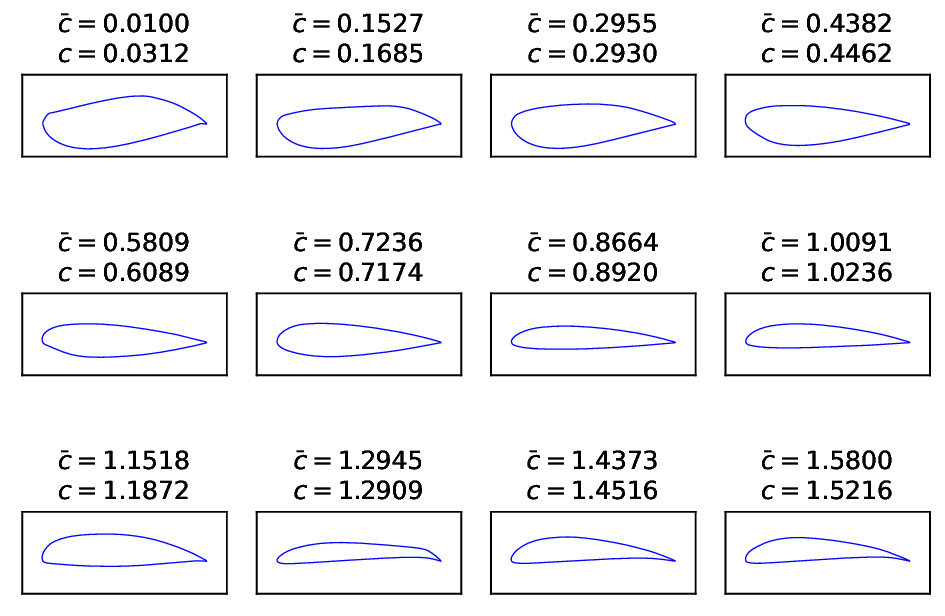}
\caption{Generated shapes by uniform intervals $\bar{c}$ with the smoothing constraint.}
\label{generated_shapes_approx_0_158_distort_continuous}
\end{figure}

\begin{figure}[!t]
\centering
\includegraphics[width=112mm]{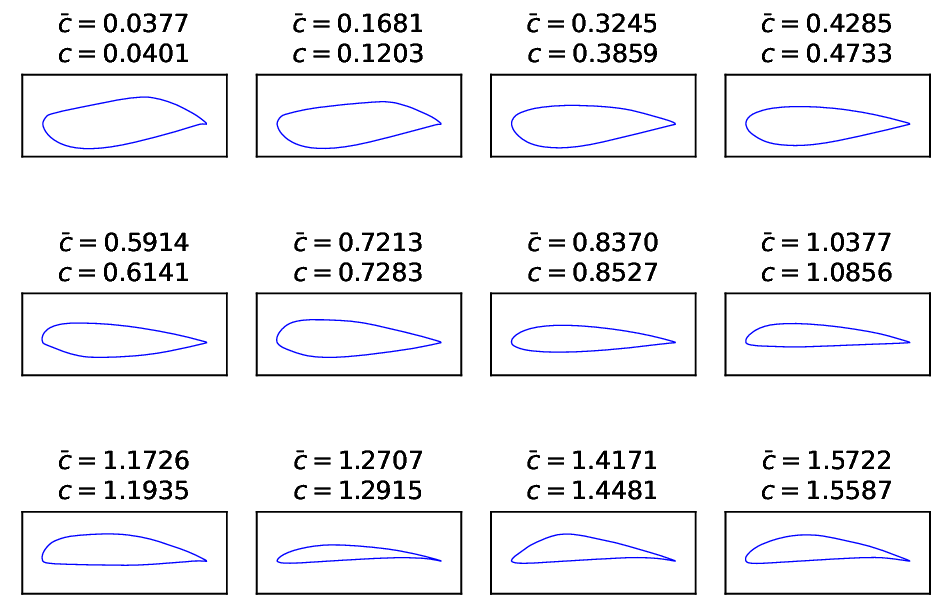}
\caption{Generated shapes by random $\bar{c}$ with the smoothing constraint.}
\label{generated_shapes_approx_0_158_distort_random}
\end{figure}

\begin{figure}[!tb]
    \centering
    
    \begin{subfigure}[b]{0.3\textwidth}
        \centering
        \includegraphics[width=\textwidth]{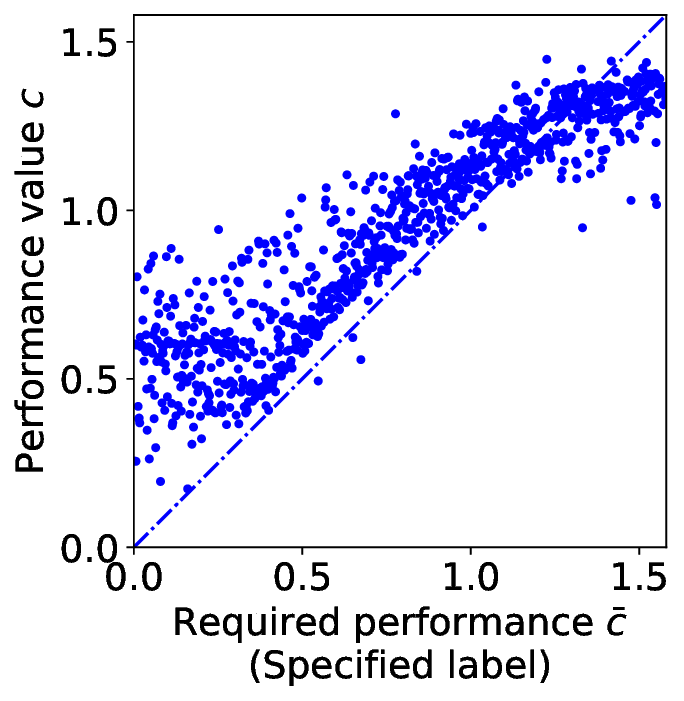}
         \caption{Scatter plot at the initiation.}
    \end{subfigure}
    \hfill
    \begin{subfigure}[b]{0.3\textwidth}
        \centering
        \includegraphics[width=\textwidth]{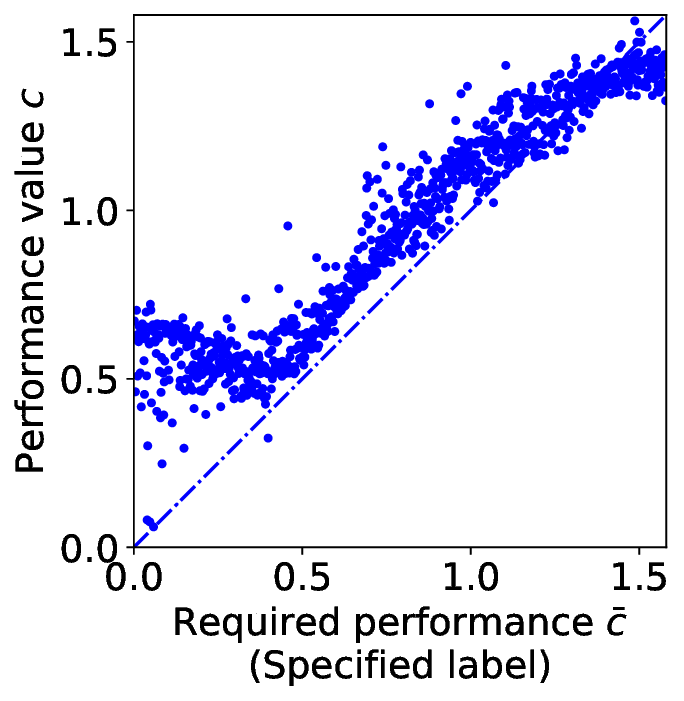}
        \caption{Scatter plot at 5000 epoch.}
    \end{subfigure}
    \hfill
    \begin{subfigure}[b]{0.3\textwidth}
        \centering
        \includegraphics[width=\textwidth]{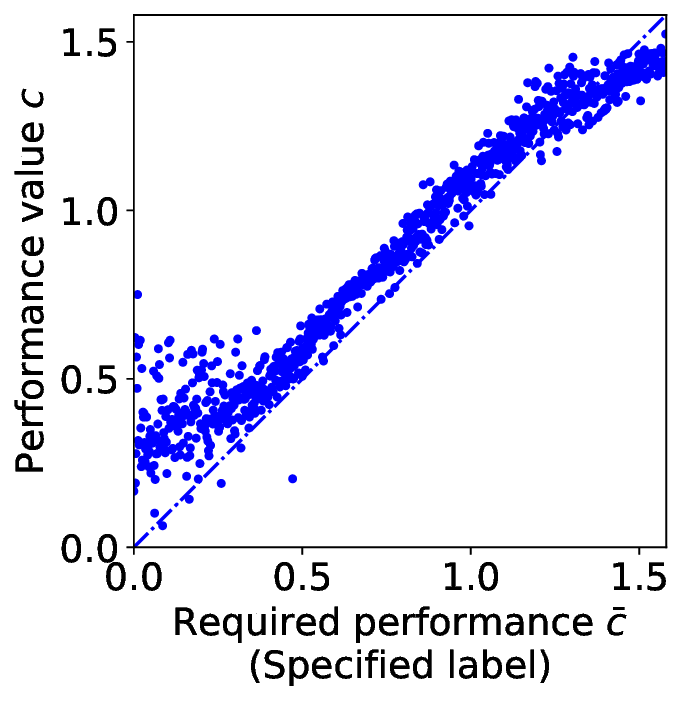}
        \caption{Scatter plot at 10000 epoch.}
    \end{subfigure}
    
    \vspace{1em}
    
    \begin{subfigure}[b]{0.3\textwidth}
        \centering
        \includegraphics[width=\textwidth]{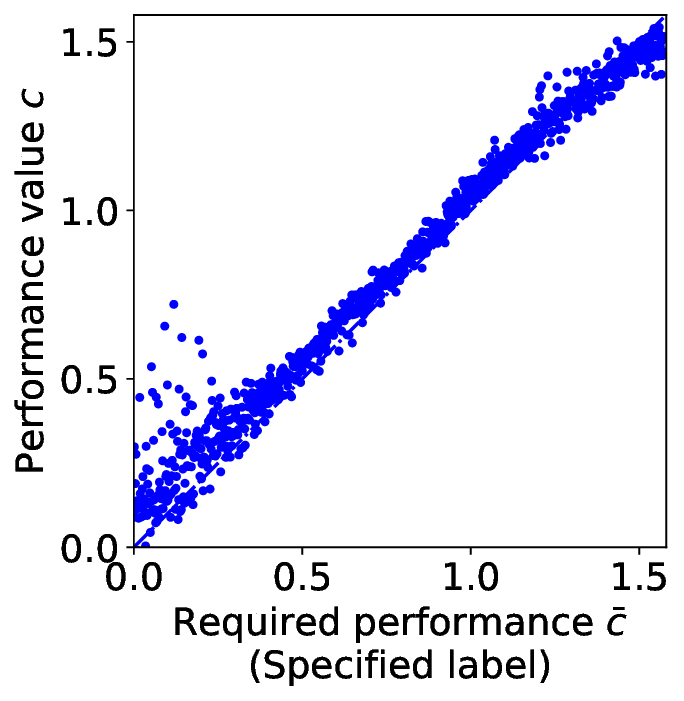}
        \caption{Scatter plot at 15000 epoch.}
    \end{subfigure}
    \hspace{0.05\textwidth}
    \begin{subfigure}[b]{0.3\textwidth}
        \centering
        \includegraphics[width=\textwidth]{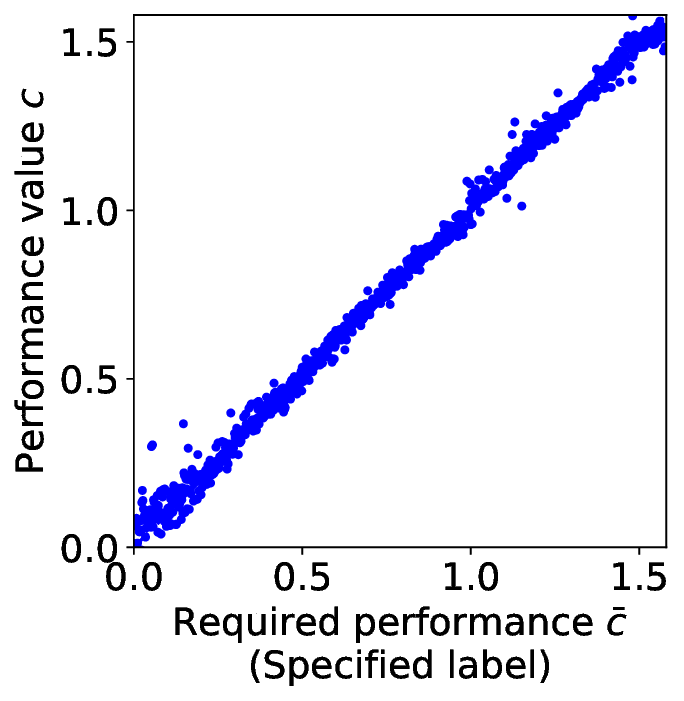}
        \caption{Scatter plot at 20000 epoch.}
    \end{subfigure}
    
  \caption{Scatter plot transition of PG-cWGAN-gp with smoothing constraint.}
  \label{fig:five_phi_scatters}
\end{figure}

\newpage
Next, we present the results generated by the model incorporating smoothing constraints.
An experiment was performed with $\lambda_{\rm phi}=1$ in Eq. \ref{V_G}. 
Fig. \ref{generated_shapes_approx_0_158_distort_continuous} and Fig. \ref{generated_shapes_approx_0_158_distort_random}  show the shapes generated with smoothing constraint, and Fig. \ref{fig:five_phi_scatters} depicts the scatter plot transition. 
Fig. \ref{fig:five_phi_scatters} shows that, as with Fig. \ref{fig:five_no_phi_scatters}, it is possible to generate shapes with progressively smaller errors as the learning progresses.

Some of the output shapes (Fig.\ref{shapes_approxi_0_158_continuous}, Fig.\ref{shapes_approxi_0_158_random}, Fig. \ref{generated_shapes_approx_0_158_distort_continuous}, and Fig. \ref{generated_shapes_approx_0_158_distort_random}) are different from the training dataset, and successfully generate completely new data. It is because the proposed model no longer uses the training dataset as true data. 
It overcomes the limitation of the GAN model in that the generative models output data is similar to the training data and  completely new data cannot be generated.

\begin{figure}[!t]
		\centering
		\includegraphics[width=140mm]{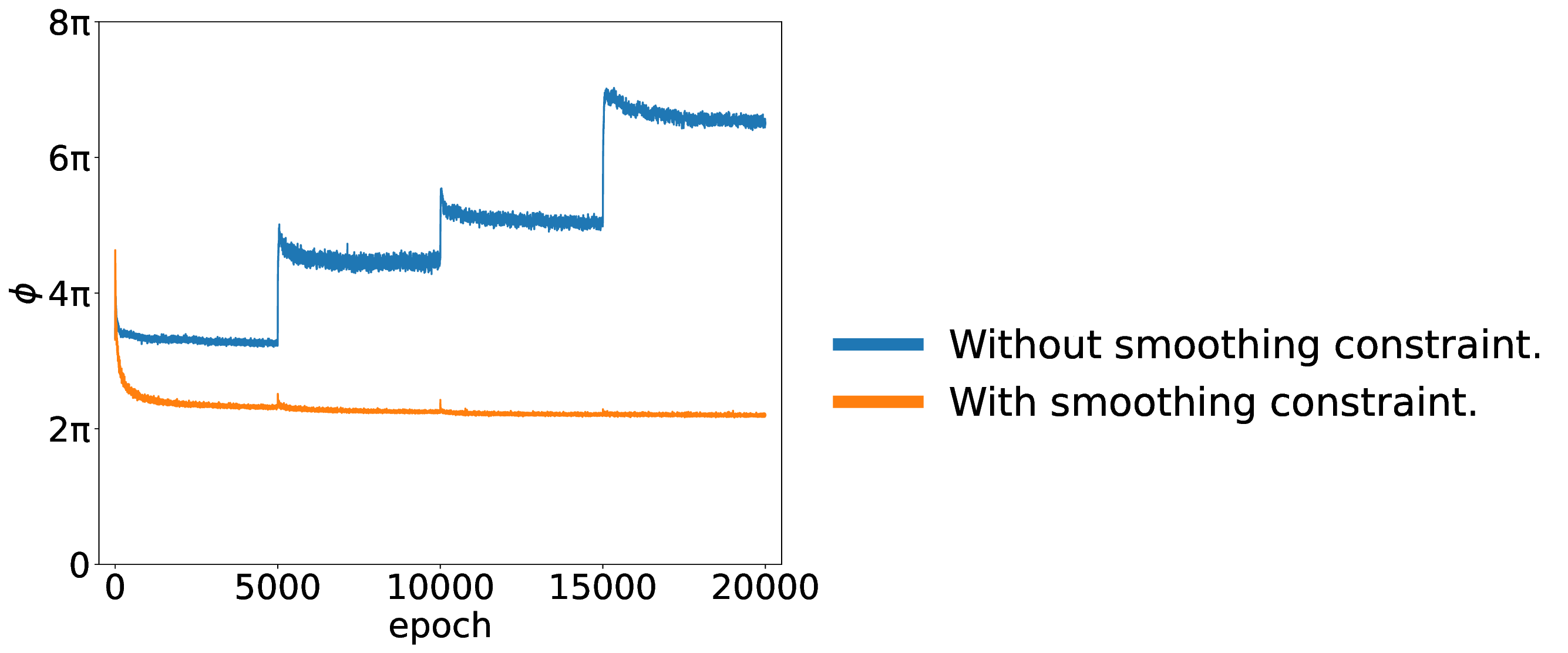}
			\caption{Transition of distortion $\phi$ for each epoch without and with the smoothing constraint.}
                \label{phi_transition}
\end{figure}

As the training progresses, the distortion degree $\phi$ increases without the smoothing constraints. 
However, in the case of smoothing constraints, the training progresses to minimize $\phi$ (Fig. \ref{phi_transition}).

\newpage
Comparison between prior studies using cGAN and cWGAN-gp and the proposed PG-cWGAN-gp model, 
based on the performance metrics of success/failure/nonconvergence ratios, MAE, and shape distortion $\phi$ (Table \ref{comparison_of_models}).
\begin{table}[!t]
\centering
\caption{Comparison of models.}
\label{comparison_of_models}
\begin{tabular}{lrrrrr} \hline
 & \multicolumn{1}{c}{success $\uparrow$} & \multicolumn{1}{c}{failure $\downarrow$} & \multicolumn{1}{c}{not converge $\downarrow$} & \multicolumn{1}{c}{MAE $\downarrow$} & \multicolumn{1}{c}{$\phi$ $\downarrow$} \\ \hline 
cGAN & 0.7 $\%$ & 65.1 $\%$ & 28.1 $\%$ & 0.147 & 5.25 $\pi$ \\
cWGAN-gp & 1.0 $\%$ & 94.0 $\%$ & 5.0 $\%$ & 0.211 & 3.42 $\pi$ \\
PG-cWGAN-gp & 61.0 $\%$ & 9.6 $\%$ & 29.4 $\%$ & 0.026 & 6.50 $\pi$ \\
PG-cWGAN-gp with smoothing constraints & \textbf{87.5 $\%$} & \textbf{8.1 $\%$} & \textbf{4.4 $\%$} & \textbf{0.022} & \textbf{2.27 $\pi$} \\ \hline
\end{tabular}
\end{table}
It was found that PG-cWGAN-gp was able to generate with higher accuracy than the previous method. In particular, 
the PG-cWGAN-gp model with smoothing constraints performed the best because it generated smooth shapes. These smooth shapes caused the XFoil calculations to converge.


\section{Conclusion}

Previous studies on shape generation approaches using deep generative models have encountered large errors in the performance values.
This is because the shapes generated by these models do not always satisfy the physical equations. To overcome this issue, the present study proposed a PG-GAN based inverse design method that used XFoil to link the lift coefficient to the generated shapes. 
The conventional PINN model requires the implementation of physical equations inside the neural networks, which prevents the use of commercial and general-purpose softwares. 
The proposed PG-GAN model enables the use of software by performing physical calculations outside the neural network. Essentially, this model requires performance calculations for all generated shapes; however, it requires a long computation time.
Therefore, we introduced an approximation model to reduce the number of calculations required. Although the results fell short of the exact algorithm, they exhibited a drastic improvement in accuracy compared to previous studies.
By describing the smoothing constraints in the model, smooth and small-error shapes were generated while linking the required performance to the performance value. 

The output of the proposed method differed from that of the training dataset and was completely new. 
It  is because the proposed method no longer uses a training dataset. 
A limitation of the generative model is that it can not produce completely new shapes. 
The proposed method overcame this problem.

\section*{Acknowledgements}
This study was supported by JSPS KAKENHI Grant Numbers JP21K14064 and JP23K13239.

\bibliographystyle{unsrt}

\end{document}